\documentclass{article} 
\usepackage{iclr2027_conference,times}

\usepackage{amsmath,amsfonts,bm}

\def\eqref#1{equation~\ref{#1}}

\def\1{\bm{1}}

\DeclareMathAlphabet{\mathsfit}{\encodingdefault}{\sfdefault}{m}{sl}
\SetMathAlphabet{\mathsfit}{bold}{\encodingdefault}{\sfdefault}{bx}{n}

\usepackage{hyperref}
\usepackage{url}
\usepackage{booktabs}
\usepackage{array}
\usepackage{multirow}
\usepackage{graphicx}
\usepackage{amsmath}
\usepackage{xcolor}
\usepackage{algorithm}
\usepackage{algorithmic}
\hypersetup{colorlinks=true,linkcolor=blue!60!black,citecolor=blue!60!black,urlcolor=blue!60!black}

\newcommand{\graft}{\textsc{Graft}}
\newcommand{\tailacc}{tail@acc}

\title{GRAFT: Restoring the Missing Sampling\\Interface in Continuous Diffusion Language\\ Models}

\author{Haoning Yang \quad Lili Pan\\
University of Electronic Science and Technology of China\\
\texttt{202521011616@std.uestc.edu.cn}, \texttt{lilipan@uestc.edu.cn}}

\iclrfinalcopy

\begin{document}

\maketitle

\begin{abstract}
Continuous diffusion language models generate a whole sequence in parallel, never drawing a single word at any intermediate step; only
the final step turns embeddings into text. Models such as ELF now
report autoregressive-level quality at a fraction of the sampling
steps taken by discrete-token diffusion models, and this design
principle---no step-wise discretization---is what their own ablations
credit for the gain. We show that this same principle carries a
hidden cost: a collapse of the lexical long tail. Measured against
held-out human text, ELF's output matches the human rate on common
words almost exactly, yet its rarest words appear at only $0.30\times$
the human rate, a gap that does not close from 105M to 652M
parameters and that no sampler setting these models expose can reach.
A same-size GPT-2, in contrast, spans zero to above the human rate by
its decoding rule alone. We trace the difference to what continuous
diffusion models lack: with no step that ever draws a discrete word,
every step is instead driven by an average over the candidates still
in play, and averaging a rare candidate together with the common word
competing for its slot lands on the common word. An autoregressive
model avoids this because its decoding rule samples one candidate
rather than averaging over all of them, so the problem never arises
there. We propose \textbf{GRAFT}, a training-free method that grafts
this missing choice back onto a frozen model: while the model is
still undecided, GRAFT commits the rare words the average would
otherwise discard, then regenerates the sequence from scratch with
those commitments held in place. Across three ELF scales and on
LangFlow, an independently developed continuous dLLM, GRAFT restores
the rare-word share of generated text from $0.42$--$0.77\times$ the
human rate to $0.82$--$1.10\times$, a relative gain of $34$--$109\%$, and moves the entire
word-frequency profile closer to human on every model tested, by
$2.7\times$ on ELF-B and $12\times$ on LangFlow.
\end{abstract}

\section{Introduction}
\label{sec:intro}

\begin{figure}[t]
\centering
\includegraphics[width=0.9\linewidth]
{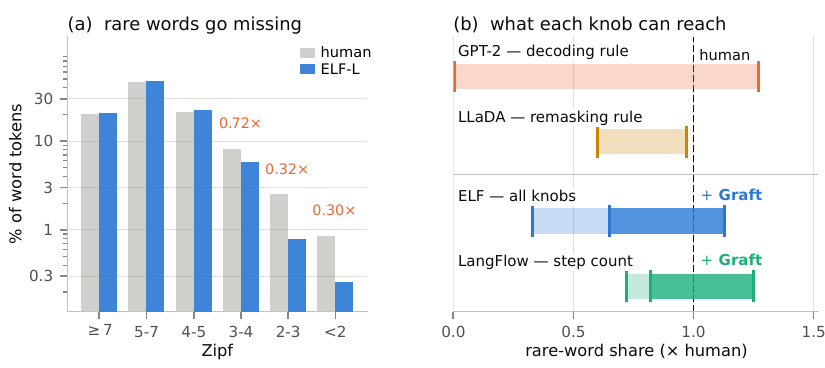}
\caption{Lexical-tail coverage depends on whether a model samples discrete words during generation. \textbf{(a)} ELF-L matches human usage for common words but substantially underproduces rare ones, reaching only $0.72\times$, $0.32\times$, and $0.30\times$ of the human rate across the three rarest Zipf bands. \textbf{(b)} GPT-2 and LLaDA, which sample discrete words during decoding, reach or approach the human rare-word rate, while ELF and LangFlow remain below it. \graft{} restores the missing tail, raising ELF and LangFlow to $1.13\times$ and $1.25\times$ of the human rate (three-seed means, Appendix~\ref{app:sweep}).}
\label{fig:fig1}
\end{figure}

Diffusion language models (dLLMs) generate all positions in parallel
and refine them over a fixed number of steps, trading autoregression's
$O(L)$ serial depth for $O(\text{steps})$ \citep{austin2021d3pm,
sahoo2024mdlm, nie2025llada}. 
dLLMs fall into two lines, distinguished by whether the diffusion process runs
over discrete tokens \citep{lou2024sedd, nie2025llada, llada2} or over continuous embeddings \citep{li2022diffusionlm, dieleman2022cdcd}.
Continuous diffusion language models historically lagged behind their discrete counterparts. Earlier approaches typically relied on either projecting the continuous state back to the token space at every denoising step \citep{li2022diffusionlm, dieleman2022cdcd}, or performing diffusion in the latent space of a separately trained autoencoder \citep{lovelace2023latent}. ELF \citep{hu2026elf} removes both requirements: the entire denoising trajectory remains in the token-embedding space, and discretization is performed only once, at the final step. This design substantially improves efficiency, allowing ELF to outperform discrete diffusion baselines with only 32 sampling steps instead of 1024 and with roughly one-tenth of the training tokens. LangFlow \citep{chen2026langflow} independently achieves comparable performance through a similar continuous-space formulation. Ablation studies in both works attribute much of these gains to the same key design choice: eliminating intermediate, step-wise discretization.

However, the same design choice that makes continuous dLLMs efficient
also introduces a hidden lexical cost: collapse of the vocabulary tail
(Figure~\ref{fig:fig1}a). Here, rarity is defined by corpus frequency
rather than model probability; proper names, technical terms, and other
low-frequency words together account for more than a tenth of human
text. On held-out text from ELF's training distribution, common words
appear at nearly the human rate, while the rarest occur at only
$0.30\times$ that rate. This gap persists from 105M to 652M parameters
(Table~\ref{tab:buckets}) and is not closed by sampler tuning.
When a rare word is lost, it is typically replaced by a more common word
or a fabricated \emph{non-word}, reducing lexical fidelity. For example,
one ELF summary reads ``New Barcelona president Neyll has resigned,''
producing a fluent sentence with an invented name
(\S\ref{sec:collapse}). Yet such failures are easily missed by standard
metrics: GPT-2-based generative
perplexity and unigram entropy. The former can favor text that avoids
rare words, while the latter is too coarse to reveal a deficit
concentrated in the lexical tail.

The cause lies in a missing step in the continuous dLLM pipeline.
Autoregressive models first predict a vocabulary distribution and then
apply a decoding rule---such as greedy, top-$k$, or nucleus sampling---
to select a discrete word. We refer to this distribution-to-word
operation as a \emph{sampling interface}. Continuous dLLMs lack such an
interface before the final step: their denoising head instead predicts
continuous embeddings by regression throughout the trajectory, while
the decoding head turns them into words only at the end. Because regression
is driven toward the conditional mean, competing candidates are averaged
rather than sampled; when a rare word competes with a more probable
common word, the resulting embedding tends to move toward the common
one. By the time the decoding head is finally read, each position has
already narrowed to only $1.02$--$1.05$ effective candidates, leaving
little room for temperature scaling, sampling, or beam search to recover
the lost tail. In short, rare-word probability survives inside the
model, but the generation process never turns it into a discrete choice.

\begin{center}
\emph{Can this missing sampling interface be restored at inference time,
without retraining the model?}
\end{center}

We answer yes with \textbf{GRAFT}, a training-free method that restores
a discrete choice while rare-word alternatives are still present.
Part-way through generation, GRAFT reads the decoding head and identifies
rare words that retain substantial probability but are losing to more
common competitors. It then restarts generation from noise with these
words fixed from the first step, allowing the remaining sequence to form
around them. Repeating this read-and-restart procedure accumulates a set
of committed rare words, after which a final generation produces the
output. Sub-word constraints and an anti-repetition correction
\citep{lpir2026} mitigate the artifacts introduced by fixing words.

Our contributions are threefold. First, we identify a lexical long-tail
collapse in continuous dLLMs, showing that the effect persists across
model scales and sampler settings when measured against a
same-distribution human baseline (\S\ref{sec:collapse}). Second, we
uncover its mechanism: rare words are lost when a more common competing
word accumulates sufficient probability, and the competitor's probability
alone strongly predicts which rare words will be suppressed
(\S\ref{sec:cause}). Third, we introduce \textbf{GRAFT}, a training-free
method that restores the missing lexical tail. Across three ELF scales
and an independent LangFlow implementation, GRAFT raises the rare-word
share to $0.82$--$1.10\times$ the human rate, moves the overall
word-frequency distribution closer to human---by $2.7\times$ on ELF-B
and $12\times$ on LangFlow---and brings ELF-L's $\zeta$ to within
$0.010$ of the human value
(\S\ref{sec:results}--\S\ref{sec:ablation}).

\section{Related work}
\label{sec:related}

Diffusion language models fall into two lines, distinguished by
whether the diffusion process runs over discrete tokens or over
continuous embeddings.

The discrete line began with D3PM, which first carried diffusion to
discrete tokens, using an absorbing mask state as the corruption
process \citep{austin2021d3pm}; MDLM then reduced masked diffusion to
a weighted cross-entropy that trains as easily as a language model
\citep{sahoo2024mdlm}; and LLaDA scaled the recipe to 8B parameters and
autoregressive-level quality \citep{nie2025llada}. Every model in this
line draws a token from a categorical distribution at each unmasking
step, so remasking order and temperature come as decoding knobs for
free.

The continuous line diffuses in embedding space instead. Diffusion-LM
opened it, but had to round the embeddings back to the vocabulary at
every step to continue \citep{li2022diffusionlm}; the variants that
followed kept some per-step anchor to the vocabulary, and none caught
up with autoregression at equal compute. ELF removed these anchors
entirely, keeping the whole trajectory in embedding space and
discretizing only at the last step, and reached autoregressive-level
quality in 32 steps \citep{hu2026elf}; LangFlow reached the same point
independently, by a related route \citep{chen2026langflow}. The
anchors this line removed, however, were also its only place to draw a
word. This paper measures what that removal costs, and puts the draw
back before generation begins.

We list here only the most representative work in each line; a fuller
review is given in Appendix~\ref{app:related}.

\section{Preliminaries}
\label{sec:prelim}

\textbf{The models.} We study ELF \citep{hu2026elf} and LangFlow
\citep{chen2026langflow}, two independently developed continuous dLLMs.
ELF represents a sentence as $x\in\mathbb{R}^{L\times d}$ and learns a
flow from noise to data. Under \emph{x-prediction}, given a noised state
$z_t$ at time $t$, the model predicts the clean embedding
$\hat{x}(z_t,t)$ over ${\sim}32$ denoising steps. A single network has
two output heads: the denoising head, trained with MSE, drives every
step of the trajectory, while the decoding head, trained with
token-level cross-entropy through a shared unembedding, is used only at
the final step. LangFlow (${\sim}256$ steps) has only the decoding head and
feeds back its expected embedding at every step.

\textbf{Word-frequency metrics, measured against human text.}
We compare model outputs with a human reference to measure the
lexical-tail deficit and the overall word-frequency distribution.
\emph{Zipf frequency} is the log-scale corpus frequency of a word
\citep{vanheuven2014zipf}, with larger values indicating more common
words. The \emph{tail} is the fraction of word tokens with Zipf ${<}4$,
further split into $[3,4)$, $[2,3)$, and ${<}2$ to measure the depth of
the deficit. A \emph{non-word} is a string with no lexicon entry
\citep{lpir2026}, which may replace a lost rare word. \emph{Unigram
entropy} measures lexical diversity and helps show why standard
diversity checks can miss a tail-specific collapse. A sample's
\emph{Zipf profile} is the distribution of lexicon words across six
Zipf bands (Appendix~\ref{app:protocol}). We compare this profile with
the human reference using \emph{JSD}, where $0$ indicates an exact
match. Finally, the \emph{Zipf coefficient} $\zeta$ is the magnitude
of the slope fitted to the rank--frequency curve; larger $\zeta$
indicates stronger concentration on a small set of common words.

\section{The collapse: rare words missing, no metric noticing}
\label{sec:collapse}

\begin{table}[t]
\centering
\small
\renewcommand{\arraystretch}{0.92}
\caption{Word-frequency profile of unconditional generation (1{,}000
samples each) against held-out human text from the training
distribution. High-frequency buckets match human almost exactly; the
deficit is confined to the tail, and the rarest bucket does not move
from 105M to 652M parameters.}
\label{tab:buckets}

\resizebox{0.8\linewidth}{!}{%
\begin{tabular}{lrrrrr}
\toprule
Zipf bucket (\% of word tokens) & ELF-B & ELF-M & ELF-L & Human & ELF-L / Human \\
\midrule
$\geq 7$            & 20.60 & 20.62 & 21.02 & 20.38 & $1.03\times$ \\
$[5,7)$             & 53.07 & 51.59 & 48.13 & 46.20 & $1.04\times$ \\
$[4,5)$             & 20.81 & 20.69 & 22.64 & 21.27 & $1.06\times$ \\
$[3,4)$             &  3.14 &  4.93 &  5.82 &  8.12 & $\mathbf{0.72\times}$ \\
$[2,3)$             &  0.44 &  0.68 &  0.80 &  2.51 & $\mathbf{0.32\times}$ \\
$<2$                &  0.27 &  0.24 &  0.26 &  0.86 & $\mathbf{0.30\times}$ \\
non-word (no entry) &  1.67 &  1.25 &  1.35 &  0.66 & $\mathbf{2.05\times}$ \\
\midrule
types / 100k        & 2{,}949 & 3{,}288 & 3{,}505 & 5{,}329 & $0.66\times$ \\
\bottomrule
\end{tabular}%
}
\end{table}

This section quantifies the lexical-tail collapse introduced above.
Table~\ref{tab:buckets} groups word tokens from $1{,}000$
unconditional samples at each ELF scale by Zipf frequency and compares
them with held-out human text from the same corpus. The problem is not
that the text is broadly poor, but that ELF overuses common words and
underuses rare ones: common-word statistics closely match human text,
while the deficit is concentrated in the tail. The missing mass shifts
partly to mid-frequency words and partly to non-words, whose rate is
about $2\times$ the human level. This raises three questions: whether
the deficit shrinks with scale or tuning, whether it depends on the
task, and whether standard metrics would detect it.

The deficit does not disappear with scale or sampler tuning. It remains
as the model grows from 105M to 652M parameters, and none of the sampler
knobs we test closes the gap. Its severity also depends on how much rare
content must be generated rather than copied. Translation, which can
copy rare tokens from the source, is largely unaffected. Summarization,
which must generate names, fabricates proper nouns absent from both
source and dictionary at $3.53\%$, compared with $0.82\%$ in human
references. Unconditional generation, where all content is
self-produced, is the most affected. Details of this task ordering are
given in Appendix~\ref{app:protocol}.

Two simpler explanations can be ruled out before turning to the
mechanism: undertraining and insufficient sampler tuning. Undertraining
cannot explain the effect, since ELF-B saw $5\times$ as many training
tokens as GPT-2 small at comparable scale. Nor is the issue simply the
choice of decoding setting. The same GPT-2 checkpoint moves from
$0.00\times$ to $1.27\times$ the human tail by changing only its
decoding rule, whereas ELF's full sweep over steps, guidance,
temperature, and model scale reaches only $0.33$--$0.65\times$
(Figure~\ref{fig:fig1}b; Appendix~\ref{app:sweep}). The key difference
is therefore not training scale, but whether the model exposes a
word-level sampling interface.

Standard metrics also fail to reveal the problem. GPT-2-based
generative perplexity can favor text that avoids rare words because
GPT-2 assigns them lower probability. On ELF-B, perplexity
\emph{improves} from $44.5$ to $19.0$ as the number of steps increases
from 16 to 64, even while $4$-gram repetition rises $12\times$
(Appendix~\ref{app:sweep}). Unigram entropy is similarly insensitive:
it appears normal even though ELF's Zipf slope is $-1.227$, compared
with $-0.986$ for human text, because the statistic is dominated by
common words (\S\ref{sec:prelim}). Thus, a deficit concentrated in the
lexical tail can remain hidden under standard evaluations, extending to
the lexical axis what \citet{zheng2025masked} and \citet{lpir2026}
observed for low-diversity text.

\section{Root cause: the average inside the denoising head}
\label{sec:cause}

The lexical tail disappears from the output, yet no sampler setting
recovers it (\S\ref{sec:collapse}). We trace this failure to the
denoising head itself: because it is trained by regression, it predicts
an average over competing word candidates, and that average is pulled
toward a more probable common word. Three findings support this
mechanism; Appendix~\ref{app:exclusions} rules out eight alternative
explanations.

\begin{figure}[t]
\centering
\includegraphics[width=0.9\linewidth]{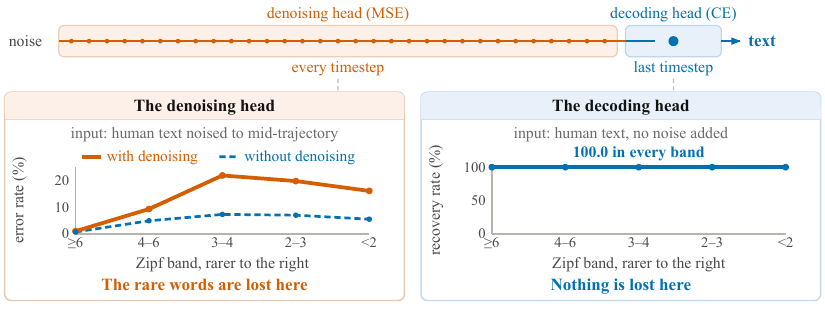}%
\caption{Where rare words are lost. ELF uses a denoising head (MSE) at
every timestep and a decoding head (CE) only at the final readout.
\textbf{Right:} on clean held-out human embeddings, the decoding head
recovers all words across $124{,}212$ positions (Finding~1).
\textbf{Left:} on noised text at an intermediate generation level,
errors increase sharply only after denoising and primarily on rare words:
on ELF-B, the gap grows from $1.0\%$ vs.\ $0.7\%$ in the commonest band
to $21.9\%$ vs.\ $7.3\%$ in $[3,4)$ (Finding~2).}
\label{fig:heads}
\end{figure}

\textbf{Finding 1: The rare words are already gone before the decoding
head reads them.}
The decoding head maps embeddings to words, so it is the natural first
suspect. If it were biased toward common words, it could itself produce
the observed tail collapse. We test this using embeddings that still
contain genuine human text: $128$ held-out OpenWebText documents are
encoded with ELF's frozen T5 encoder and fed into ELF-B without added
noise, using the same readout as at the end of generation. Across
$124{,}212$ positions, the decoding head recovers $100.00\%$ of the
original words (Table~\ref{tab:appexcl}, Appendix~\ref{app:exclusions}).
Thus, whenever a rare word is still present in the embedding, the
decoding head can recover it. The tail must therefore already be absent
from the embeddings that reach the final readout.

\textbf{Finding 2: The rare words are lost inside the denoising head,
and specifically because they are rare.}
With the final decoder ruled out, we next examine the denoising head,
which drives every generation step. We noise $96$ held-out OpenWebText
documents to a level encountered part-way through generation and feed
the same input to both heads. Because the heads share the same underlying
network and receive the same information, their training objectives are
the main difference (Appendix~\ref{app:exclusions}). Across roughly
$75{,}000$ positions and all three ELF scales, the denoising head is
wrong $1.0$--$1.9\times$ as often as the decoding head on common words,
but $2$--$2.9\times$ as often on the rarest words
(Table~\ref{tab:apptwoheads}, Appendix~\ref{app:heads}). Its excess error is concentrated in the lexical tail,
showing that rare words are lost inside the denoising process.

\textbf{Finding 3: The rare words are lost when a more common
competitor already carries enough probability.}
If denoising averages over candidate words, then the probability of the
strongest competitor should predict whether a rare word survives. We
test this on $200$ held-out OpenWebText documents, noised as above.
For each position, decoding the noised embedding directly gives $q$,
the probability assigned to the strongest competing word; applying the
denoising head first and then decoding reveals which word survives.
Among $4{,}036$ rare positions on ELF-B, only $7\%$ lose their word in
the lowest-$q$ decile, compared with $92\%$ in the highest-$q$ decile
(Appendix~\ref{app:qdecile}). Moreover, when a rare word is lost, that
same competitor replaces it $31.8\%$ of the time. The strongest
competitor's probability therefore predicts both whether a rare word is
lost and, often, which word replaces it.

\section{\graft{}: commit the rare words the average would lose}
\label{sec:method}

\graft{} follows from \S\ref{sec:cause}: the model loses a rare word by averaging it away during generation, so we find the rare words it is about to lose and hand them back to it as given before it generates again. A word is handed over through the model's own conditioning input, and we call handing one over in this way \emph{committing} it. ELF accepts known words at arbitrary positions: it starts the generation with each given word already in place, holds it there through every denoising step, and forms the rest of the sequence around it, as it does around a prompt. LangFlow has no such input, and we emulate one by writing the given word's embedding into its position at every step (Appendix~\ref{app:protocol}). The procedure has four steps, of which the first three repeat.

\textbf{Step 1: generate, and stop part-way.} A round starts a
generation from noise, the same in every round, and runs it only as far as a point we call
the \emph{read time}; the rest of that trajectory is never run. The
read time is one denoising step inside the short window where
positions still hold several candidates: earlier, the sequence has too
little shape for those candidates to mean anything; later, each
position has already narrowed to the $1.02$--$1.05$ candidates of
\S\ref{sec:intro} (where this window sits in each model:
Appendix~\ref{app:commitlog}).

\textbf{Step 2: read the decoding head, once.} At that single point,
the decoding head---the only part of the model that names words at
all (\S\ref{sec:prelim})---ranks the whole vocabulary at every
position, and \graft{} reads out this full ranking. This is the only
point before the end of a generation at which a word is chosen.

\begin{figure}[t]
\centering
\includegraphics[width=0.9\linewidth]{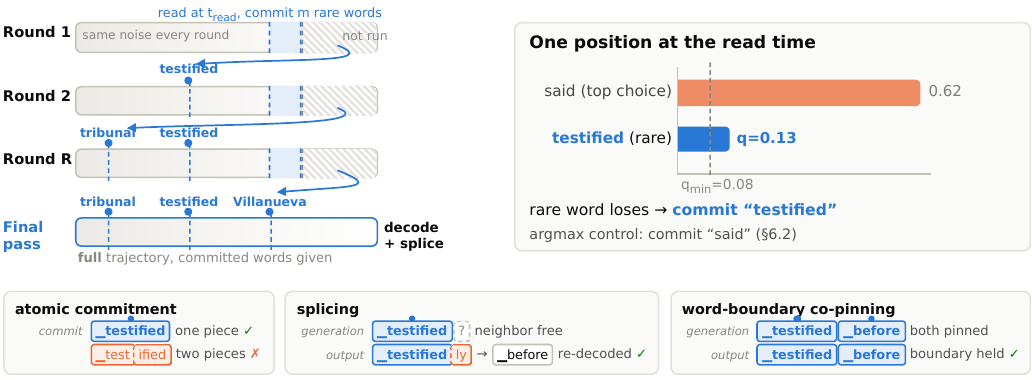}%
\caption{\graft{} commits rare words from the first denoising step and
uses restarts to preserve them. \textbf{Left:} each of the $R$ rounds
starts from the same noise with all previously committed words fixed,
runs only to the read time, and commits $m$ additional rare words; a
final full trajectory then generates the output with all commitments
held in place. \textbf{Top right:} at the read time, \graft{} commits a
rare word that retains substantial probability but loses to the top
choice, while the argmax control commits that top choice instead.
\textbf{Bottom:} three rules prevent committed words from creating
non-words.}
\label{fig:method}
\end{figure}

\textbf{Step 3: select the rare words that are losing, and commit
them.} A position qualifies when its strongest rare candidate---among
the $512$ most likely words there---carries probability above
$q_{\min}{=}0.08$ but is not the top choice; a rare word that is
already winning needs no rescue. A qualifying position must also pass
the rules of below. Among all qualifying positions,
\graft{} ranks them by that probability and commits the $m$ highest,
each to its own strongest rare candidate.

\textbf{Step 4: restart, and finish on the last round.} The generation
of Step~1 is then abandoned rather than finished. The next round
returns to that noise, with the new commitments added to the given
words, and Steps~1 through~3 run again; after $R$ rounds, a final
generation---with all committed words now given---runs to completion
and produces the text (Figure~\ref{fig:method}; Algorithm~\ref{alg:graft}
in Appendix~\ref{app:protocol}). We use $R{=}4$ and $m{=}8$, so a
sample carries up to $32$ committed words in total; that total is what
we call the \emph{dose}. This costs $R$ partial generations plus one
full one---between two and three full trajectories on ELF, depending
on where its sampler places the denoising steps---and, with overhead,
$2.2\times$ the wall clock of the model's own stock sampler
(Appendix~\ref{app:buildup}).

\graft{}'s intervention is not entirely free of side effects, and each
requires a targeted remedy.

Committing words can disrupt nearby text. Although \graft{} commits
whole words, the model operates on sub-word pieces, and a word may span
multiple positions. A commitment fixes only the position it occupies,
while neighboring positions are still generated freely. If the
following piece was originally meant to complete the displaced word,
the two may merge into a non-word. We prevent this with three rules.
First, \emph{atomic commitment} restricts commitments to single-piece
words placed only at positions that already contain a single-piece word,
avoiding partial overwrites. Second, \emph{splicing} re-decodes the
position immediately after a commitment at output and requires it to
begin a new word, preventing leftover fragments from attaching to the
committed word. Third, \emph{word-boundary co-pinning} fixes that
following position to a word-beginning piece from the model's own draft
throughout generation, so the model knows from the first step that a new
word begins there.

The lexicon used to define ``rare'' introduces a second issue. It may
contain strings that are not valid English words, as well as words that
are rare in English but common in another language, either of which
could artificially inflate the measured tail. We therefore apply two
gates. A candidate may be committed only if it is \emph{attested} in a
held-out slice of human text, and it is rejected if it is more than
$2.5$ Zipf points more common in any of thirteen other languages than
in English (Appendices~\ref{app:foreign} and~\ref{app:buckets}). These
gates affect only which words \graft{} may commit, not how the metric
counts them; in Table~\ref{tab:main}, the attestation gate is on for
both models and the language gate for ELF.

A third side effect is increased repetition. Unlike the local sub-word
artifacts above, this effect is distributed across the generated text.
It matches the self-conditioning feedback identified by
\citet{lpir2026}, where repetition accumulates along a dominant
direction in the state fed back to the model. We apply their correction
at $\lambda{=}0.5$, counteracting this direction at each denoising step.
The same direction appears in both models, while larger doses sharply
degrade generation quality (Appendix~\ref{app:negative}).

These remedies leave one design question: why must \graft{} restart from
noise rather than intervene on an ongoing trajectory? We tested two
cheaper alternatives, and both fail. The first discretizes the sequence
at a single intermediate denoising step and continues generation from
the resulting words. Early in the trajectory, positions still support
several candidates and are not yet mutually coherent, so independently
selected sub-word pieces do not compose reliably; later, when
discretization becomes safe, the text has already largely settled and
the intervention has little effect. The second alternative applies a
softer intervention over several steps, repeatedly moving selected
positions toward the rare word's embedding. Small moves are undone by
subsequent denoising and the sampler returns to its original output.
Strong moves can preserve the target word, but sharply increase
non-words and local corruption. In our experiments, only a word supplied
before the first denoising step persists reliably, and only in this
regime does the identity of the committed word matter. This motivates
\graft{}'s restart-and-commit design; the corresponding measurements
are reported in Appendix~\ref{app:negative}.

\section{Experiments}
\label{sec:setup}

We test whether committing the rare words that would otherwise be lost
restores the lexical tail. We first describe the models, metrics, and
dose setting, then report the main results, metric-validity checks,
alternative explanations, and ablations.

We evaluate ELF-B/M/L (105M/342M/652M official checkpoints, 32 sampling
steps) and LangFlow (652M, 256 steps), all frozen, with unconditional
generation and $n{=}200$ samples per arm unless noted. Both model
generation and the human reference come from OpenWebText, ELF's training
corpus, at the same $1024$-token sequence length. Thus, all statistics
are measured against a same-distribution human baseline. We do not
report generative perplexity because, as shown in
\S\ref{sec:collapse}, it rewards the lexical deficit we measure.

Table~\ref{tab:main} reports six metrics. \emph{Tail\,\%} is the share
of rare word tokens, and \emph{$\times$human} expresses it relative to
the human rate of $11.88\%$. \emph{Non-word\,\%} controls for
fabricated strings: a tail gain is meaningful only if it is not obtained
by increasing non-words, whose human rate is $1.14\%$
(Appendix~\ref{app:protocol}).

To measure the full word-frequency distribution, we report
\emph{JSD} and \emph{$|\Delta\zeta|$}. JSD compares the complete Zipf
profile with the human reference, while $|\Delta\zeta|$ measures the
difference in Zipf concentration from the human value
$\zeta{=}0.971$; lower is better for both. Finally, \emph{Types} counts
distinct generated words and provides a general measure of vocabulary
breadth.

\subsection{Setting the dose}
\label{sec:dose}

The dose---the number of words \graft{} commits---controls a trade-off:
too few commitments leave the tail short, while too many overshoot the
human tail and distort the overall frequency profile
(Figure~\ref{fig:dose}, Appendix~\ref{app:lfdose}). We use up to
$m{=}8$ qualifying positions per round for $R{=}4$ rounds at all three
ELF scales. This single untuned setting places ELF-B just below the
human tail, ELF-M near it, and ELF-L slightly above it.

The same dose overshoots on LangFlow because its stock sampler already
reaches $0.77\times$ the human tail, compared with $0.42\times$ for
ELF-B. We therefore select a LangFlow-specific dose using JSD on one
seed and evaluate the selected setting on two held-out seeds. The chosen
dose commits none from $[3,4)$ and at most one each from $[2,3)$ and
below per round, bringing LangFlow to $1.03\times$ the human tail and
the closest frequency profile in Table~\ref{tab:main}. We use this
held-out selection only where the fixed dose overshoots; all ELF results
remain untuned (Appendix~\ref{app:heldout}).

\subsection{Results}
\label{sec:results}

\begin{table}[t]
\centering
\small
\renewcommand{\arraystretch}{0.92}
\caption{Main results on \emph{unfiltered} output ($n{=}200$ per arm).
\graft{} rows and the stock rows of ELF-L and LangFlow average three
seeds; the remaining rows are single arms. The LangFlow \emph{tuned}
row uses the held-out dose of \S\ref{sec:dose}; no ELF arm is tuned.
\emph{Best-of-4} is the equal-budget control of \S\ref{sec:controls}.
Bold marks the value closest to human within each model.}
\label{tab:main}

\resizebox{0.81\linewidth}{!}{%
\begin{tabular}{lrrrrrr}
\toprule
 & \multicolumn{3}{c}{lexical tail, and its control}
 & \multicolumn{3}{c}{distance to human profile} \\
\cmidrule(lr){2-4}\cmidrule(l){5-7}
 & tail\,\% & $\times$human & non-word\,\%
 & JSD$\,\downarrow$ & $|\Delta\zeta|\,\downarrow$ & types \\
\midrule

\textsc{Human reference}
& 11.88 & 1.00 & 1.14 & --- & --- & 17\,726 \\

\midrule
\multicolumn{7}{@{}l}{\textbf{ELF-B}~\textnormal{\footnotesize(105M)}}\\
\quad Stock
& 5.01 & 0.42 & 2.64 & .0134 & .182 & 11\,795 \\
\quad Stock, best-of-4
& 5.97 & 0.50 & 1.79 & .0101 & .175 & 11\,482 \\
\quad Argmax control
& 5.31 & 0.45 & 2.42 & .0133 & .209 & 12\,124 \\
\quad \graft{}
& \textbf{9.76} & \textbf{0.82} & 2.57
& \textbf{.0049} & \textbf{.130} & \textbf{13\,829} \\

\midrule
\multicolumn{7}{@{}l}{\textbf{ELF-M}~\textnormal{\footnotesize(342M)}}\\
\quad Stock
& 5.75 & 0.48 & 1.97 & .0090 & .138 & 14\,111 \\
\quad \graft{}
& \textbf{12.02} & \textbf{1.01} & 3.11
& \textbf{.0079} & \textbf{.066} & \textbf{15\,695} \\

\midrule
\multicolumn{7}{@{}l}{\textbf{ELF-L}~\textnormal{\footnotesize(652M)}}\\
\quad Stock
& 7.65 & 0.64 & 2.67 & .0052 & .058 & 15\,478 \\
\quad \graft{}
& \textbf{13.05} & \textbf{1.10} & 2.88
& \textbf{.0046} & \textbf{.010} & \textbf{17\,106} \\

\midrule
\multicolumn{7}{@{}l}{\textbf{LangFlow}~\textnormal{\footnotesize(652M, independent implementation)}}\\
\quad Stock
& 9.16 & 0.77 & 3.28 & .0025 & \textbf{.011} & 18\,896 \\
\quad Argmax control
& 9.75 & 0.82 & 3.47 & .0059 & .104 & \textbf{18\,572} \\
\quad \graft{}
& 14.81 & 1.25 & 3.58 & .0058 & .063 & 18\,981 \\
\quad \graft{} (tuned)
& \textbf{12.23} & \textbf{1.03} & 3.79
& \textbf{.0002} & .045 & 18\,940 \\

\bottomrule
\end{tabular}%
}
\end{table}

\graft{} restores the lexical tail without simply creating fabricated
words (Table~\ref{tab:main}, Figure~\ref{fig:fig1}b). Across the four
models, the rare-word share rises from $0.42$--$0.77\times$ the human
rate to $0.82$--$1.10\times$, gains of $3.1$--$6.3$ percentage points.
Meanwhile, non-word rates rise by at most $1.1$ points and even fall on
ELF-B ($2.64\to2.57$), indicating that the added vocabulary is largely
recovered rather than fabricated.
The broader frequency profile improves as well. JSD to the human
profile decreases on all four models, by $2.7\times$ on ELF-B and
$12\times$ on LangFlow, while ELF-L's $\zeta$ comes within $0.010$ of
the human value. LangFlow is the one exception on $\zeta$: its stock
sampler is already within $.011$ of human, and no \graft{} dose improves
on it, so we make the $\zeta$ claim only for ELF
(Appendix~\ref{app:lfdose}).

\subsection{Metric validity}
\label{sec:premises}

For tail\,\% to be a valid measure of lexical reach, rather than an
easily gamed number, two premises need to hold, and a further check
confirms the resulting gains are not an artifact of sample quality.

The first premise is that a rare word counted in the tail is used
\emph{appropriately}, not merely inserted: GPT-2-large, used nowhere
in the sampler, rates the committed words no less apt in context than
rare words occurring naturally in human text
(Appendix~\ref{app:buckets}). The second is that a word is only
counted as rare if it is genuinely rare in English, not merely rare
because it belongs to another language: every tail figure is
discounted for words that are rare in English but common elsewhere
(Appendix~\ref{app:foreign}).

As a further check on validity, Appendix~\ref{app:accept} re-reads
every arm after discarding samples whose non-word or $4$-gram
repetition rate exceeds the human sample's own 95th percentile,
following \citet{zheng2025masked} and \citet{lpir2026}. The gains
reported below survive this filter, confirming they are not driven by
a handful of degenerate samples. We note, however, that no system,
ours included, closes the gap to the human non-word rate ($2.0$--$3.8\%$
against $1.14\%$).

\subsection{Alternative explanations}
\label{sec:controls}

The title names what this section rules out: two alternative
explanations, besides the identity of the committed word, that could
account for the gains reported above.

The first is that the gain comes from the act of committing rather
than from the word committed: restarting, holding words fixed, and
the sub-word repairs might each lift the tail on their own, regardless
of which word is chosen. To separate the two, on ELF-B and LangFlow we
run an argmax control that shares the \graft{} arm's positions,
mechanics, dose, and compute, but commits the model's top choice at
each position instead of the rare word. This control is null on every
column of Table~\ref{tab:main}: on ELF-B the tail moves only from
$5.01$ to $5.31$, and $\zeta$ moves in the wrong direction ($.209$
against $.182$); on both models, the bootstrap interval of its tail
gain includes zero (Appendix~\ref{app:accept}). The gain therefore
rests on which word is committed, not on the act of committing.

The second alternative is compute: because the method resembles
best-of-$K$ selection, the gain could simply reflect spending more
compute per sample. The four restarts of a \graft{} sample run only to
the read time, and its measured wall clock is $2.2\times$ the stock
sampler's, so we compare against the equal-budget baseline: the best
of four stock samples. At that budget, \graft{} still leads on the
unfiltered tail: best-of-4 reaches $0.50\times$ the human tail against
$0.82\times$ on ELF-B (Table~\ref{tab:main}), and $0.86\times$ against
the tuned $1.03\times$ on LangFlow (Appendix~\ref{app:heads}), a lead
the tuned dose loses on the accepted set (Appendix~\ref{app:accept}). Nor does
this gap close with still more compute: the stock sampler catches up
on LangFlow only at best-of-16, and on ELF-B not at all
(Appendix~\ref{app:heads}).

\subsection{Ablation study}
\label{sec:ablation}

Table~\ref{tab:ablation} adds the three commitment rules of
\S\ref{sec:method} to the stock ELF-B sampler one at a time, to
separate two questions the rules could in principle answer: where the
tail gain comes from, and what the rules themselves are for.
The first row answers the first question. Committing rare words with
no rule at all already lifts the tail from $4.54$ to $7.92$---three
quarters of the eventual gain---so the tail comes from the act of
commitment itself, not from any of the three rules. This bare
commitment is not free, however: the non-word rate rises from $2.35$
to $4.14$, and $13\%$ of the committed words reach the output as some
other string rather than the word intended.
The remaining rows answer the second question: the three rules exist
to repair that fabrication, not to grow the tail further. Atomic
commitment and splicing, added together, bring every committed word
through to the output intact, taking fidelity to $100\%$.
Word-boundary co-pinning then drives the non-word rate down further,
to $1.78$---below the stock sampler's own $2.35$---at a small cost of
$0.5$ points of tail.

In sum, the tail comes from committing the rare word; the three rules
are not where it comes from, but they are what bring fabrication down,
in the end below the stock sampler's own rate. The two gates and the
anti-repetition correction are ablated separately in
Appendices~\ref{app:buildup} and~\ref{app:buckets}.

\begin{table}[t]
\centering
\small
\renewcommand{\arraystretch}{0.92}
\caption{Ablation of the three commitment rules on ELF-B. Each row adds
one rule to the previous setting; \emph{Fidelity} is the fraction of
committed words preserved in the final output. The last row also changes
the sampler's step distribution, so its $0.5$-point tail reduction
reflects both effects. The full \graft{} setting further includes the
two gates and the anti-repetition correction.}
\label{tab:ablation}
\begin{tabular}{lrrr}
\toprule
 & tail\,\% & non-word\,\% & fidelity\,\% \\
\midrule
stock                                    & 4.54 & 2.35 & --- \\
\quad $+$ rare commitment, no rule       & 7.92 & 4.14 & 86.7 \\
\quad $+$ atomic commitment              & 8.86 & 2.71 & 87.4 \\
\quad $+$ splicing                       & 9.61 & 2.47 & 100.0 \\
\quad $+$ word-boundary co-pinning       & 9.10 & 1.78 & 100.0 \\
\bottomrule
\end{tabular}
\end{table}

\section{Limitations and conclusion}
\label{sec:limitations}

We show that continuous diffusion language models suffer from a lexical
long-tail collapse caused by the absence of an intermediate sampling
interface: rare words remain represented while positions are uncertain,
but are progressively washed out by conditional-mean denoising before
the final decoding step. \graft{} restores this missing interface at
inference time by identifying losing rare-word candidates, committing
them, and regenerating around those commitments. Without any training,
it raises the rare-word share from $0.42$--$0.77\times$ to
$0.82$--$1.10\times$ the human rate across three ELF scales and an
independent implementation, at $2.2\times$ the wall-clock cost. Several
directions remain open, including training-time mechanisms that preserve
rare-word mass without restarts, richer use of the decoding distribution,
extension to broader conditional tasks and data distributions, and a
formal analysis of conditional-mean readout in embedding space together
with human evaluation.

\subsubsection*{Reproducibility statement}
All models are public checkpoints (ELF-B/M/L, LangFlow OWT); the human baseline is held-out OpenWebText. The protocol (bars, bootstrap, clean-PPL) is specified in \S\ref{sec:setup} and Appendix~\ref{app:protocol}, which also gives \graft{}'s exact procedure as Algorithm~\ref{alg:graft}; defaults are in \S\ref{sec:method}; probe code, per-arm generation outputs, per-commitment paired logs (position, committed token, counterfactual token, probabilities, entropy, final token), the human reference sample and the statistics implementations will be released publicly on publication.

\subsubsection*{Ethics statement}
This work diagnoses and mitigates a text-generation defect; it does not introduce new data collection or human subjects. Two considerations. First, the defect itself has a safety face: models that fabricate proper nouns at $4\times$ the human rate produce plausible-looking misinformation (invented people, places, organizations); restoring real rare words reduces one fabrication channel but does not verify factuality---\graft{} commits words the model supports, not words that are true. The attestation gate of \S\ref{sec:method}, on in every arm of Table~\ref{tab:main}, commits only words \emph{attested in held-out human text}, which removes the channel by which the repair could itself invent proper nouns ($2.1\%$ of ELF and $0.8\%$ of LangFlow commitments before the gate, $0\%$ after); measured on XSum, \graft{} has no detectable effect on unsupported-entity rates in either direction (Appendix~\ref{app:xsum}). Second, increasing rare-word coverage can surface offensive or private strings memorized from the training corpus at a higher rate; our probes measured no verbatim-memorization pathology in these models (Appendix~\ref{app:negative}), but deployments should keep standard output filtering in place.

\subsubsection*{AI use statement}
Generative AI tools were used substantially in this work, in the following required-disclosure categories: assisting with the design of experiments and providing feedback on methods (probe and control-arm design, dose and window selection); implementing methods (probe and sampler code); supporting analysis and interpreting results (statistical readouts, mechanism synthesis); and drafting and editing this manuscript, including its structure. Experiment execution was supervised and all judgments of experimental validity, retraction decisions, and final claims were reviewed and verified by the authors. AI tools were not used to generate synthetic training data, propose mathematical propositions or proofs, or fabricate any reported measurement. The authors take full responsibility for the content of this paper.

\bibliography{references}
\bibliographystyle{iclr2027_conference}

\appendix
\section{Protocol details}
\label{app:protocol}

This appendix specifies how \graft{} runs and how its outputs are measured; quantities used by a single analysis, such as those of the mechanism probes, are defined where that analysis is reported. We first give the procedure (Appendix~\ref{app:procedure}) and the metrics together with the human reference they are read against (Appendix~\ref{app:metrics}), and then the configurations and statistics used across the appendices (Appendix~\ref{app:configs}). The last two parts report the measurements behind the task ordering of \S\ref{sec:collapse} (Appendix~\ref{app:tasks}) and four checks that the measurements are not artifacts of the tail metric, the acceptance bars of \S\ref{sec:premises}, the GPT-2-large appropriateness judge of \S\ref{sec:premises}, or the sampler (Appendix~\ref{app:checks}).

\subsection{The \graft{} procedure}
\label{app:procedure}

Algorithm~\ref{alg:graft} states the procedure. The paragraphs below define its notation, give the settings of Table~\ref{tab:main}, and describe how the procedure is implemented on each model.

\begin{algorithm}[ht]
\caption{\graft{} for one sample on a frozen checkpoint. Notation and the settings of Table~\ref{tab:main} are given below.}
\label{alg:graft}
\begin{algorithmic}[1]
\STATE $C \leftarrow \emptyset$
\FOR{round $r = 1 \dots R$}
  \STATE run the flow from the same fixed noise in every round, \textbf{up to $t_{\mathrm{read}}$ only}, with every $(i, v)\in C$ given as a condition at every step
  \STATE at $t_{\mathrm{read}}$: read the decoding head's distributions $p_i$ and the current draft $\hat{w}$ (defined below)
  \STATE $\mathcal{E} \leftarrow \{\,i \text{ not in } C:\ q_{\mathrm{rare}}(i)>q_{\min},\ \mathrm{rare}(i)\neq\arg\max p_i,\ \mathrm{atomic}(i)\,\}$
  \STATE in decreasing order of $q_{\mathrm{rare}}$, add $(i,\mathrm{rare}(i))$ to $C$ for up to $m$ positions $i\in\mathcal{E}$, each at least $s$ tokens from every position already in $C$
  \STATE word-boundary co-pinning: for each new $i$ whose right neighbor $\hat{w}_{i+1}$ is not a word-continuation piece, also add $(i{+}1,\hat{w}_{i+1})$ to $C$
\ENDFOR
\STATE run the \textbf{full} trajectory from the same noise with all of $C$ given
\STATE decode; write each committed token back; re-decode any word-continuation piece right after a committed position, excluding such pieces
\end{algorithmic}
\end{algorithm}

\textbf{Notation.} $C$ is the set of commitments, each a pair $(i,v)$ of a position $i$ and a token $v$. For each position $i$, the decoding head gives a distribution $p_i$ over the vocabulary. Among the $512$ most likely words of $p_i$, $\mathrm{rare}(i)$ is the most probable rare word that the gates allow, where a rare word is one with Zipf frequency $0{<}z{<}4$ (Appendix~\ref{app:metrics}), and $q_{\mathrm{rare}}(i)$ is its probability. The gates are the two gates of \S\ref{sec:method}: the attestation gate, on both models, and the \emph{language gate}, on ELF only, which rejects a candidate that is more than $2.5$ Zipf points more common in any of thirteen other languages than in English (Appendix~\ref{app:foreign}). Because they remove candidates before $\mathrm{rare}(i)$ is chosen, a top choice $\arg\max p_i$ that is itself a rare word but fails a gate does not block its position: the next allowed rare candidate can still be committed. The \emph{draft} $\hat{w}$ is the text that the run would produce at the read time. On LangFlow, it is $\arg\max p_i$; on ELF, it is the decoding of the denoising head's current prediction of the clean embedding $\hat{x}$, the same prediction that is fed back for self-conditioning after the anti-repetition correction, so it need not coincide with $\arg\max p_i$. Finally, $\mathrm{atomic}(i)$ holds when the draft has a single-piece word at $i$ and $\mathrm{rare}(i)$ is a single-piece word, and $\mathcal{E}$ is the set of qualifying positions of \S\ref{sec:method}. A \emph{word-continuation piece} is a sub-word token that continues the preceding word rather than beginning a new one.

\textbf{Settings.} Table~\ref{tab:main} uses $R{=}4$ rounds, $m{=}8$ commitments per round, $q_{\min}{=}0.08$, and a spacing of $s{=}4$ tokens, with word-boundary co-pinning (\emph{co-pinning} for short) always on; on ELF-B only, the sampler's step distribution is also moved toward later steps, a change we call the \emph{step-schedule shift} (Appendix~\ref{app:configs}). The read time is $t_{\mathrm{read}}{=}0.20$ on ELF, whose time runs from noise at $0$ to data at $1$, and $0.72$ on LangFlow, whose time runs the other way. Every round and the final trajectory reuse the same seed, so their initial noise and per-step noise are identical. What changes between rounds is the conditioning: the set of committed words and, on ELF, their encoding, which is recomputed from each round's draft. As a result, the conditioning on ELF can change between rounds even when no new word is committed, since existing conditions, such as the source article in the summarization experiment of Appendix~\ref{app:xsum}, are re-encoded together with the new draft.

\textbf{How commitments are given.} ELF receives committed words through its native conditioning input. After each round, the draft, with every committed word substituted at its position, is encoded by the T5 encoder with all positions attended, and the encodings at the committed positions---and at any other given positions, such as a source text---become the conditions for the next round. LangFlow has no conditioning input, so we hold the committed positions fixed instead: at every step, the noised state $z_t$, the clean-embedding prediction, and the self-conditioning input are all set to the committed word's embedding at those positions.

\textbf{Correction and dose variants.} At every step, the anti-repetition correction of \citet{lpir2026} ($\lambda{=}0.5$) subtracts $\lambda\mathbf{u}$ from the self-conditioning input, where $\mathbf{u}$ is the repetition direction of \S\ref{sec:method}. On ELF, $\mathbf{u}$ is normalized to unit length; on LangFlow, it is an unnormalized difference of means, so the same $\lambda$ is not the same correction strength on the two models. A tuned dose, as in Table~\ref{tab:main}'s LangFlow row (\S\ref{sec:dose}), replaces the top-$m$ selection with a \emph{band quota}, a number of commitments per round in each of the bands $[3,4)$, $[2,3)$, and below $2$, written as a triple: $(0,1,1)$ on LangFlow and, in the appendices, $(8,5,3)$ on ELF, both at $q_{\min}{=}0.02$. Its commitment count is therefore not bounded by $Rm$: on seed $31337$, the ELF-B tuned arm makes $8{,}793$ commitments per $200$ samples, or about $44$ per sample.

\textbf{The argmax control.} The argmax control of \S\ref{sec:controls} qualifies and ranks positions by the same rule as \graft{}, so it commits only where a rare candidate is losing. Its committed positions nonetheless differ from those of the \graft{} arm, for two reasons: its atomic check applies to the word it actually commits, and its drafts diverge from the \graft{} arm's after the first round.

\textbf{LangFlow implementation.} LangFlow uses the GPT-2 BPE vocabulary and samples with $256$ deterministic Euler steps on a linear time grid from $1-10^{-5}$ to $10^{-5}$, which the model's own schedule maps to noise levels. At every step, its prediction of the clean embedding is the average of the token embeddings weighted by the decoding head's distribution, which is the average of \S\ref{sec:cause} in explicit form. LangFlow's sampler thus has no separate denoising head: it computes the decoding head's distribution at every step and averages over it, and only at the final step does it turn that distribution into discrete words. On LangFlow, the read time $t_{\mathrm{read}}{=}0.72$ falls at about step $72$ of $256$, and the draft is the top choice at every position. The anti-repetition correction acts only on the self-conditioning input; its direction is estimated on LangFlow itself and left unnormalized, and the committed positions are fixed again after it is applied. The attestation gate uses the same \emph{attestation lexicon} as on ELF, the words of documents $1{,}000$--$30{,}999$ of the OpenWebText shard behind the human reference, and no language gate is applied. Table~\ref{tab:apphyper} collects these settings.

\subsection{Metrics and the human reference}
\label{app:metrics}

Every word-frequency statistic in this paper is computed in the same way on model output and on human text. Two tokenization rules are used, however, and the human reference takes a different value under each (Table~\ref{tab:apphuman}).

\textbf{Tokenization.} Both rules start from whitespace tokens, strip the surrounding punctuation, and fold case. The \emph{alphabetic} rule then keeps only alphabetic strings of length at least $2$; it defines the Zipf buckets of Table~\ref{tab:buckets}, the acceptance bars and accepted-set tail defined below, the JSD, and Table~\ref{tab:ablation}. The \emph{all-token} rule instead keeps every non-empty token, including numbers and hyphenated or apostrophe forms, and defines the tail, non-word, $\zeta$, and types columns of Table~\ref{tab:main}. This is why the human non-word rate reads $0.66\%$ under the first rule and $1.14\%$ under the second.

\textbf{Tail and non-words.} Each counted token is looked up in the \texttt{wordfreq} English lexicon \citep{speer2022wordfreq}, which gives its Zipf frequency $z$; a Zipf frequency of $2$ corresponds to about $0.1$ occurrences per million tokens. The tail of \S\ref{sec:prelim} counts words with $0{<}z{<}4$; non-words, which have no entry and so $z{=}0$, are counted separately. In Table~\ref{tab:main}, the numerator of the tail also leaves out \emph{foreign-dominant} words, those more than $2.5$ Zipf points more common in one of the thirteen languages of the language gate than in English (Appendix~\ref{app:foreign}); we call this the \emph{foreign discount}. The denominator, like the non-word, $\zeta$, and types columns, keeps every token.

\textbf{JSD.} The JSD columns of Tables~\ref{tab:main}, \ref{tab:appheldout}, and~\ref{tab:applfdose} compare Zipf profiles of the alphabetic words of a text over six bands, $z{\geq}6$, $[5,6)$, $[4,5)$, $[3,4)$, $[2,3)$, and $0{<}z{<}2$. Non-words and foreign-dominant words are left out, the band shares are renormalized, and the JSD uses log base~$2$. These six bands differ from the Zipf buckets of Table~\ref{tab:buckets}.

\textbf{Types and repetition.} \emph{Types/100k} is the number of distinct counted words per $100$k counted tokens, pooled over an arm. The types column of Table~\ref{tab:main} is instead a raw count, which grows with sample size, so we read it at a common size: each arm pools its $n{=}200$ samples, and the human entry is the mean over $20$ random draws of $200$ reference documents ($17{,}726\pm247$, compared with $45{,}430$ when all $1{,}000$ documents are read). Repetition is measured as the \emph{4-gram repetition} of \S\ref{sec:premises}, the standard repeated-4-gram rate over whitespace tokens, $100\cdot\sum_\nu(\mathrm{count}(\nu)-1)/\#\text{4-grams}$, with $\nu$ running over the distinct 4-grams.

\textbf{Zipf slope and $\zeta$.} The Zipf slope of a sample is the least-squares slope of log frequency on log rank over its $5{,}000$ most frequent types. Human English sits near $-1$. The slopes quoted in \S\ref{sec:collapse}, $-1.227$ for ELF-B and $-0.986$ for the human sample, come from the scoring behind Table~\ref{tab:buckets}, which tokenizes with the regular expression \texttt{[A-Za-z]+}; this is close to the alphabetic rule but keeps one-letter words and splits at apostrophes. The Zipf coefficient $\zeta$ of Table~\ref{tab:main} comes from the same fit under the all-token rule. On that tokenization, the human sample gives $\zeta{=}0.971$ and the ELF-B batch of Table~\ref{tab:buckets} gives $1.188$, so the two recipes differ only in tokenization. Finally, $|\Delta\zeta|$ is computed per seed and averaged where an arm has several seeds.

\textbf{The human reference.} The human reference consists of the first $1{,}000$ documents of the first OpenWebText shard, cut to the same $1024$-token length as the generations so that per-sample rates are comparable. Between $996$ and $1{,}000$ of them are scored: the all-token scoring of Table~\ref{tab:main} keeps all of them, whereas the accept-then-measure scoring, defined below, drops any text, human or generated, with fewer than $20$ alphabetic words, which removes four documents. Their Zipf slope of $-0.986$ matches that of human English. The reference is same-distribution text rather than unseen text: ELF trains on its full OpenWebText split, which has no validation part and contains this shard, so these documents are part of ELF's training data; we did not check LangFlow's training split. Where this paper calls the reference \emph{held-out}, it means only that the reference is disjoint from the documents that build the attestation lexicon, documents $1{,}000$--$30{,}999$ of the same shard (Appendix~\ref{app:buckets}). The three mechanism probes of \S\ref{sec:cause} read the first $128$, $96$, and $200$ of the reference documents.

\textbf{Acceptance bars.} We call the re-reading of \S\ref{sec:premises} the \emph{accept-then-measure} protocol. It accepts a sample only if it clears bars set at the human sample's own per-sample 95th percentiles: a non-word rate of at most $2.737\%$ and, for the dual bar, a 4-gram repetition of at most $10.12\%$. A sample that clears the non-word bar passes the \emph{single bar}, one that clears both passes the \emph{dual bar}, and \tailacc{}, the \emph{accepted-set tail}, is the tail share of the accepted samples. The share of samples accepted is the \emph{accept rate} (\emph{accept} for short). Under the dual bar, the human sample accepts $90.5\%$ of its own documents, with an accepted-set tail of $11.17\%$; this is the human reference point of every accept-then-measure reading in the appendices. Table~\ref{tab:main} instead compares against the unfiltered $11.88\%$ of the all-token rule.

\subsection{Configurations, runs, and statistics}
\label{app:configs}

\textbf{Configuration names.} Besides the terms of the main text, the appendices use a few short names. \graft{} is the configuration of Table~\ref{tab:main}, with the settings of Appendix~\ref{app:procedure}. The \emph{champion} is an earlier configuration with rare commitment, atomic commitment, splicing, and the anti-repetition correction, but without co-pinning or gates, and \emph{champion$+$co-pinning} adds co-pinning, still without gates. The fixed dose of \S\ref{sec:dose} is $m{=}8$ at $q_{\min}{=}0.08$, whereas a tuned dose is a band quota (Appendix~\ref{app:heldout}). On LangFlow, the tuned dose is $(0,1,1)$, selected by a held-out procedure. On ELF, it is $(8,5,3)$, which was compared with three other quotas on ELF-B only and reused on ELF-M and ELF-L, so it was not selected per model. Table~\ref{tab:main}'s statement that no ELF arm is tuned refers to its own rows; in the appendices, \emph{tuned} names the band-quota dose on either model. \emph{Gated} marks arms with the gates on. The \emph{ladder} is a per-sample controller that tries a fixed list of configurations from strongest to weakest and keeps the first output that passes both bars (Appendix~\ref{app:buildup}); it is not part of \graft{} and enters no result of the main text. Finally, an \emph{off-manifold} breakdown is the abrupt collapse of the text, with exploding repetition or non-words, that bounds the usable strength of a correction \citep{lpir2026}.

\textbf{The step-schedule shift.} On ELF-B, the champion, champion$+$co-pinning, and the \graft{} and argmax-control rows of Table~\ref{tab:main} also use the step-schedule shift, which changes the sampler's step distribution by moving the center of ELF's logit-normal step schedule from its default $-1.5$ to $-1.0$. This moves the median step from $t{\approx}0.18$ to $t{\approx}0.27$, closer to where rare words are decided (Appendix~\ref{app:commitlog}). The shift does not itself buy tail: run alone on ELF-B without commitment (seed $31337$), it lowers the unfiltered tail from $4.54$ to $3.99$ and the non-word rate from $2.35$ to $2.01$, both under the alphabetic rule. ELF-M, ELF-L, and every stock row use the default schedule, and LangFlow's step schedule has no such parameter.

\textbf{Which runs the appendices read.} The \graft{} rows of Table~\ref{tab:main} are gated runs averaged over three seeds. Much of the appendices instead reads earlier configurations, which we keep because each isolates one component or answers one question. Readings labeled champion or champion$+$co-pinning come from ungated runs, many of them on a single seed, and the ladder rows are ungated as well. Each table caption states the configuration, gates, and seeds of its rows, and Appendix~\ref{app:repro} gives the settings and per-seed values of Table~\ref{tab:main} itself. Four analyses address questions outside Table~\ref{tab:main}, all in Appendix~\ref{app:heads}: the comparison with Marginal-Conditioned Bridges \citep[MCB;][]{mcb2026} and INCO \citep{inco2026}, the two closest samplers proposed for continuous-space language models; the LLaDA control behind the LLaDA range of Figure~\ref{fig:fig1}b; prefix-conditioned generation; and the re-scoring of the \emph{appropriateness judge}, the GPT-2-large rating of committed words in \S\ref{sec:premises}, in the draft from which each word was selected (Appendix~\ref{app:checks}).

\textbf{Two sampling pipelines on ELF.} The ELF measurements come from two pipelines that load different weights. ELF's own generation script decodes with each checkpoint's exponential-moving-average (EMA) weights; it produces the ELF columns of Table~\ref{tab:buckets}, the step-and-guidance sweep of Table~\ref{tab:appsweep}, and the translation and summarization counts of Appendix~\ref{app:tasks}. Our probe pipeline, which implements \graft{}, instead loads the training weights stored in the same checkpoints; it produces every ELF arm of Table~\ref{tab:main}, every other \graft{} comparison on ELF, and the mechanism probes of Appendix~\ref{app:exclusions}. The two sets of weights differ, and stock ELF-B reads differently in the two pipelines at the same sampler settings: under the alphabetic rule, ELF's script gives a tail of $3.85\%$ with $1.66\%$ non-words ($1{,}000$ samples), whereas the four stock seeds behind the ELF-B best-of-4 row of Table~\ref{tab:main} give $4.37$--$6.12\%$ with $2.32$--$2.81\%$ non-words. Every \graft{} gain on ELF is measured against a stock run of the probe pipeline, so the two arms of a comparison always share their weights. Figure~\ref{fig:fig1}b, which places the sweep beside the \graft{} arms, puts readings from both pipelines on one axis (Appendix~\ref{app:sweep}).

\textbf{Bootstrap.} Deltas against the stock sampler use a nonparametric bootstrap over samples, with $1{,}000$--$2{,}000$ resamples. In each replicate, the arm's and the stock run's sample sets are resampled with replacement, acceptance is re-applied within the replicate---so the accepted set is not frozen---and the accepted-set tail difference is recorded. We report the percentile $95\%$ CI and a two-sided sign-based $p$, and we write $p{<}0.001$ when no replicate falls on the other side of zero. With $1{,}000$ resamples, however, the smallest $p$ that the bootstrap can resolve is $0.002$, so for such readings the notation means only that $p$ is below the bootstrap's resolution. Where two arms share generation noise, as a \graft{} arm and its argmax control do when they run in the same batches with the same seeds, the pairing happens at generation time, and the bootstrap itself remains two-sample.

Four exceptions apply. First, the two best-of-16 comparisons of Appendix~\ref{app:heads} resample sample indices jointly for both arms, which makes them a paired bootstrap. Second, the foreign-discounted $\Delta$ of Tables~\ref{tab:appforeign} and~\ref{tab:appaccept}, and the discounted control intervals of Appendix~\ref{app:foreign}, resample within each arm's accepted set, so acceptance is not re-applied in those replicates. Third, the summarization comparison of Appendix~\ref{app:xsum} pairs the two arms by source article and resamples articles jointly. Fourth, the unfiltered readings of Appendix~\ref{app:accept} recompute the tail of Table~\ref{tab:main} in every replicate and apply no acceptance filter. For multi-seed rows (Tables~\ref{tab:main} and~\ref{tab:appaccept}), we average the per-seed readings; in Table~\ref{tab:appaccept}, each seed's $\Delta$ is taken against the same stock run. Seed-level ranges are given where noted, as in Table~\ref{tab:appscale}.

\textbf{Clean-PPL.} The appendices price each component by \emph{clean-PPL}, the generative perplexity that GPT-2-large assigns to the accepted samples only; on it, the human reference reads $16.66$. Because acceptance already caps non-words and repetition, degenerate text cannot improve this number. Generative perplexity nonetheless penalizes rare words by construction (\S\ref{sec:collapse}), so we report it as a price rather than a headline: Appendix~\ref{app:buildup} gives the price of each component, and Appendix~\ref{app:buckets} the tail bought per point of clean-PPL.

\subsection{Conditional tasks: translation and summarization}
\label{app:tasks}

Section~\ref{sec:collapse} orders three tasks by how much rare content the model must produce itself, from translation, which can carry it over from the source, to unconditional generation, which must produce all of it. This subsection gives the measurements behind the first two.

\textbf{Translation.} We measure translation with ELF's German-to-English checkpoint (ELF-B-de-en) on the first $500$ sentences of the WMT14 German--English validation set, using its ODE sampler with $64$ steps and classifier-free guidance scale~$2$ on the source; it reaches a BLEU of $30.79$. The same code scores the model output and the human reference translations. A word of four or more letters counts as untranslated German when its German Zipf frequency is at least $3$ and it either exceeds the English one by at least $2$ or has no English entry. Such words make up only $0.039\%$ of the output ($2$ of $5{,}164$ words) and none of the references. For numbers, $4.20\%$ of the source numbers are missing from the output and $7.28\%$ of the output numbers have no support in the source, against $9.79\%$ on both counts for the references. Because the rare content of a translation is already in its source and can be carried over, translation is largely unaffected by the collapse.

\textbf{Summarization.} The proper-noun fabrication rate of \S\ref{sec:collapse} is measured on XSum with ELF-B's XSum-conditioned checkpoint and ELF's own XSum sampler ($64$ ODE steps, classifier-free guidance~$2$ on the source article), against the dataset's human reference summaries. It uses a \emph{strict coined-string} criterion: a capitalized word in the generated summary counts as coined when it appears neither in the source article nor in the \texttt{wordfreq} lexicon, the dictionary of \S\ref{sec:collapse}. Under this criterion, $59$ of $1{,}671$ generated proper nouns are coined ($3.53\%$), compared with $14$ of $1{,}697$ in the references ($0.82\%$), a ratio of $4.3\times$. Although the script behind these counts was not archived, rerunning the archived extraction of capitalized words, our proper-noun candidates, with the same criterion finds the same $59$ and $14$ coined names among $1{,}749$ and $1{,}761$ proper nouns ($3.37\%$ and $0.80\%$, $4.2\times$), so the ratio stands. Under the looser criterion of counting every entity unsupported by the source, the comparison on that same extraction reverses ($25.90\%$ against $34.24\%$), because human abstractive references import world knowledge. This reversal is why we measure invention with the strict criterion. Unconditional generation, where the model must produce all of its rare content, shows the full collapse of Table~\ref{tab:buckets}.

\subsection{Checks on the metrics and the sampler}
\label{app:checks}

Four checks address the ways in which the measurements above could mislead: a loop between the selection rule and the tail metric, acceptance bars estimated on the same sample they judge, a judge that favors common words, and a sampler that silently tempers its draws.

\textbf{A tail metric without \texttt{wordfreq}.} \graft{} selects rare words with the same \texttt{wordfreq} estimates that define the tail, so a gain on that tail could partly reflect the selection rule itself. To break this loop, we rebuilt the tail metric without \texttt{wordfreq}'s frequency estimates: the \emph{top-5k list} holds the $5{,}000$ most frequent words of the human reference, and the \emph{top-5k tail} is the share of words outside that list, with \texttt{wordfreq} used only to exclude non-words. On this metric, the human reference reads $12.35$, and every arm keeps its ordering against stock. From stock to champion$+$co-pinning, the tail rises from $11.05$ to $16.48$ on LangFlow and from $6.61$ to $11.62$ on ELF-B, where the champion reads $11.26$ and the ladder $11.42$. The ordering also holds under the stricter threshold $z{<}3.5$. Under $z{<}3$, it holds on ELF-B ($0.78\to1.40$--$1.54$) but not on LangFlow, whose gain sits in $[3,4)$; there, the $z{<}3$ share barely moves, reading $1.46$ for stock, $1.41$ for the champion, $1.52$ for champion$+$co-pinning, and $1.54$ for the ladder.

\textbf{Acceptance bars on half the reference.} Because the bars are estimated from the human sample itself, we re-estimated both on two disjoint halves of that sample, which gives (non-word, repetition) bars of ($2.35\%$, $13.68\%$) and ($2.88\%$, $7.07\%$). The repetition bar is volatile, so absolute accept rates carry this uncertainty. Every ordering, however, is unchanged under either half's bars, both for each \graft{} arm against stock and for champion$+$co-pinning against champion. Reusing the bars for selection also costs little. When the four-member pool of Appendix~\ref{app:heads}---three ungated variants of \graft{}, the ladder levels of Appendix~\ref{app:buildup}, and the stock output---picks its output under one half's bars and is then scored under the bars of the whole reference, its gain falls from $+5.59$ to $+4.92$ on ELF-B and from $+5.34$ to $+4.99$ on LangFlow, that is, by only $0.3$--$0.7$ percentage points (pp).

\textbf{The appropriateness judge.} The first premise of \S\ref{sec:premises}, that committed words are used appropriately, is tested with GPT-2-large, which is used nowhere in the sampler. For each kept commitment, that is, each committed word that survives into the output, the original (\emph{v1}) judge scores the conditional log-probability of the committed word in its final context. On ELF, this scoring is word-level after T5 detokenization, over $n{=}6{,}166$ commitments of the ELF-B champion$+$co-pinning arm; on LangFlow, it is token-level, over $n{=}2{,}227$ commitments of a separate batch-2 run of the LangFlow champion (seed $42$), scored on its iterated-splicing output (Appendix~\ref{app:negative}), rather than the batch-8 champion run used elsewhere ($2{,}437$ commitments). Because raw log-probabilities favor common words through the frequency prior alone, the comparison must be matched in rarity. The reference distribution is therefore the same judge's scores at naturally occurring rare-word positions ($0{<}z{<}4$) in the human reference: $9{,}933$ positions at the word level and $11{,}514$ at the token level. We compare medians and report the fraction of commitments above the human-rare median, which is $50\%$ under perfect calibration. Both models also log the \emph{counterfactual} at each position, the top choice $\arg\max p_i$ that the argmax control would commit there, and on LangFlow we additionally report the gap between the committed word and this counterfactual.

The final context, however, was generated around the committed word. The \emph{v2} judge removes this circularity by re-scoring commitments in the \emph{selection-time draft context}, the draft from which each word was selected, recorded on a bit-identical deterministic rerun; this context contains no information about the chosen word. On ELF-B, v2 uses the same arm and unit as v1. On LangFlow, it was run on the champion$+$co-pinning arm ($2{,}429$ commitments) and scored at the word level against the $9{,}933$ word-level positions, so the two LangFlow readings differ in arm and unit as well as in context. Since draft contexts are noisier mid-trajectory text, we report v1 and v2 side by side. Not every commitment is scored: the two ELF-B judges and the LangFlow v2 judge skip a commitment with fewer than five words of context before it, a committed word that is not purely alphabetic, and a word that GPT-2 splits into more than six tokens, and the ELF-B v1 judge also skips the first three positions. The LangFlow v1 judge, which scores tokens, skips only the first position and so scores all $2{,}227$ commitments of its run. The other scored counts ($6{,}166$ and $6{,}164$ on ELF-B, and $2{,}420$ for the LangFlow v2 judge) therefore fall slightly below the arms' commitment totals ($6{,}220$ and $2{,}429$, Appendix~\ref{app:commitlog}).

\textbf{Sampler precision.} \citet{zheng2025masked} showed that float32 truncation of Gumbel noise acts as a hidden temperature drop in discrete-diffusion samplers. Our \graft{} pipeline and our \emph{clamping} pipelines, which write a decoded word into the running state mid-trajectory and hold it there as a condition (Appendix~\ref{app:negative}), sample through \texttt{torch.distributions.Categorical}, which calls \texttt{torch.multinomial}. For a single draw, this is an exponential race, $\arg\max_v p(v)/E_v$ with $E_v\sim\mathrm{Exp}(1)$, which is equivalent to the Gumbel-max trick and is likewise computed in float32, so we measured it directly. On a synthetic Zipf distribution matched to the working point of the \emph{read window}, the short window of \S\ref{sec:method} in which positions still hold several candidates (an entropy of $3.2$ nats and a top-1 probability of $0.38$; Appendix~\ref{app:commitlog}), its tail mass over $4\times10^5$ draws is $0.993\pm0.011$ of the ground truth, with no visible truncation artifact. The LLaDA control (Table~\ref{tab:appllada}) and the MCB baseline (Table~\ref{tab:appmcb}) instead add hand-written float32 Gumbel noise. On the same synthetic distribution, such a sampler reads $0.969\pm0.048$ over $2\times10^4$ draws, which is not distinguishable from $1$ at that sample size, although we did not test those two pipelines themselves. The negative results of Appendices~\ref{app:training} and~\ref{app:negative}, in which sampled and argmax words perform alike, are therefore not an artifact of our sampler.

\section{Root-cause measurements and excluded explanations}
\label{app:exclusions}

This appendix details the measurements behind \S\ref{sec:cause}. It describes the tests behind Findings~1 and~2 (Finding~3 is tested in Appendix~\ref{app:qdecile}), rules out eight alternative explanations in Table~\ref{tab:appexcl}, and characterizes the denoising head's error.

\textbf{The decoding head (Finding~1).} Finding~1 tests the decoding head on embeddings the trajectory never reaches. In this clean-embedding test, the first $128$ documents of the human reference are encoded with the frozen T5 encoder that the model was trained against and passed to the decoding head without added noise, at $t{=}1$ and through the same call that a generation's final readout uses. Over $124{,}212$ non-padding positions, the decoding head names the document's own T5 token at all but one ($99.999\%$). Its \emph{retention}, a Zipf band's share of the decoded text divided by its share of the source text, is $1.000$ in all five bands of Table~\ref{tab:apptwoheads} ($z{\geq}6$, $[4,6)$, $[3,4)$, $[2,3)$, and $z{<}2$), so the tail comes through whole ($z{<}4$: $7.454\%$ of tokens in, $7.454\%$ out). Since nothing is lost at the readout, the tail must already be missing from what reaches it.

\textbf{The two-head comparison (Finding~2).} Finding~2 compares the two heads on the same input. Both heads read out of the same network, but each is trained to receive a noised embedding under its own noise convention. We therefore calibrate the two time arguments so that the two inputs are the same vector up to a positive scale factor (the \emph{matched input}), which leaves their information identical; the objective that trained each head is then the main remaining difference. Because the matched input need not be equally typical of each head's training inputs, we compare at four noise levels and report all of them (Appendix~\ref{app:heads}). At each level, the first $96$ documents of the human reference are noised and passed through both heads. Each head names a word at every position, the denoising head through the embedding it predicts and the decoding head through its most probable word, and a position counts as an error when the named word is not the text's word. Table~\ref{tab:apptwoheads} reports the denoising head's error rate divided by the decoding head's, per Zipf band and per scale.

\textbf{Eight alternative explanations.} Table~\ref{tab:appexcl} lists eight other explanations of the collapse. Its first row restates Finding~1, and its last two restate the controls of \S\ref{sec:collapse}.

\begin{table}[ht]
\centering
\caption{Eight alternative explanations for the collapse, each with the measurement that rules it out (ELF-B unless noted). Section~\ref{sec:cause} gives the mechanism that remains.}
\label{tab:appexcl}
\small
\begin{tabular}{>{\raggedright\arraybackslash}p{0.30\linewidth}>{\raggedright\arraybackslash}p{0.62\linewidth}}
\toprule
Alternative explanation & Measurement that rules it out \\
\midrule
The decoding head favors common words & Clean embeddings of human text are decoded at $100.00\%$, with a retention of $1.000$ in every band ($n{=}124{,}212$ positions). \\
Rare words have smaller decision cells, the regions of embedding space that decode to them & They do, but only under noise far larger than the denoising head's own error. With isotropic noise of per-dimension $\sigma{=}4$, $96.1\%$ of the rarest words flip, that is, decode to another word, compared with $55.4\%$ of the commonest; at $\sigma{\leq}1$, which covers the denoising head's own errors ($\sigma{\approx}0.3$--$0.9$), at most $0.3\%$ of any band flips. On the segment from the competitor's embedding ($\kappa{=}0$) to the true word's ($\kappa{=}1$), the flip boundary is symmetric, at $\kappa^*{=}0.146$ for rare words and $0.152$ for common ones, both far from $0.5$. \\
The output is an average of two candidates & Projected onto this segment, the denoising head's output sits at $\hat\kappa{=}0.429$ and would still decode to the true word, but the perpendicular residual is $0.65\times$ the segment's length, so the output is not on the segment. \\
The self-conditioning feedback of \citet{lpir2026} & Sweeping the self-conditioning strength $\alpha$ over $0$, $0.25$, $0.5$, $0.75$, and $1$ leaves the tail flat ($4.85$, $4.84$, $4.29$, $4.49$, and $4.72$), even at $\alpha{=}0$, where \citet{lpir2026} report generative perplexity rising to $97.5$ (ELF-B, 64 steps). \\
A single error direction, removable by a one-dimensional fix & The frequency-axis direction, the normalized difference between the mean embeddings of common and rare positions, carries only $4$--$15\%$ of the error norm, and removing it does not recover accuracy ($2.87\%\to3.12\%$). \\
The denoising head picks the commonest candidate & At $35{,}513$ contested positions, taken at $t{=}0.25$ where the decoding head is uncertain and its most common candidate is not its most probable, the denoising head chooses the most common candidate only $10.1\%$ of the time (the most probable one, $20.8\%$). Where its word is among the candidates ($47.4\%$ of these positions), its mean frequency rank is $2.69$, against $3.08$ for a random candidate: a real but weak bias. \\
A different guidance scale or step count & At $32$ steps, self-conditioning guidance, ELF's classifier-free guidance on its self-conditioning input, of $1$, $3$, and $5$ gives tails of $3.88$, $3.75$, and $4.02$. From $32$ to $64$ steps at guidance $3$, it rises from $3.75$ to $4.57$ ($1.22\times$; the 64-step run also lowers the SDE churn $\gamma$, the amount of fresh noise re-injected at each step, from $1.5$ to $1.0$). The 16-step point ($7.05$, Table~\ref{tab:appsweep}) is still only $0.61\times$ human, and its accepted samples read $3.99$ (Appendix~\ref{app:sweep}). \\
Undertraining & At comparable size and with $1/5$ of ELF-B's training data, GPT-2 spans $0.00$--$1.27\times$ the human tail by its decoding rule alone (\S\ref{sec:collapse}). \\
\bottomrule
\end{tabular}
\end{table}

\textbf{The error is targeted, not noisy.} If the denoising head's errors were merely noise of the right size, other errors of the same norm should flip words about as often. We test this by replacing the error in each prediction with another of the same norm, either an isotropic random direction or a true error borrowed from another position in the batch, and decoding the result. Such replacements leave almost every word in place: isotropic ones keep $99.9$--$100\%$ of words ($t{=}0.1$--$0.5$), and permuted true errors keep $99.25$--$100\%$ ($t{=}0.2$--$0.5$), in two separate measurements. The head's own error at $t{=}0.25$, by contrast, flips $90\%$ of the rarest words, and $96.1\%$ of those flips land on a more common word (a mean Zipf increase of $2.95$). What matters is therefore the direction of the error, not its size.

\textbf{The loss compounds.} Iterating the denoising head as a map at a fixed time $t$, $x\leftarrow D(tx+(1-t)\epsilon)$, where $D$ is one pass of that head and $\epsilon$ is fresh noise, shrinks the tail of human text, which starts at $10.67$ in this compounding probe, to $6.6$, $7.6$, $8.5$, and $9.9$ at $t{=}0.3$, $0.4$, $0.5$, and $0.6$ (two runs of $10$--$12$ iterations). The more noise each step adds, the more tail is lost. Started from the model's own text ($5.44$), the map at $t{=}0.3$ settles near the same level ($7.1$). In a separate run that starts from the model's text at $5.20$, the tail stays near its start at $t{=}0.4$ and $0.6$ ($5.2$--$5.3$ after ten iterations), whereas at $t{=}0.9$ it drifts down to $4.50$.

\textbf{No geometric amplifier upstream.} In T5-small's static token-embedding table, the two rarest bands are not squeezed into a narrow cone: their cone ratio, the length of the mean of their unit-normalized embeddings ($0$ when directions spread evenly, $1$ when they coincide), of $0.23$--$0.25$ matches the $0.23$--$0.28$ of the commoner bands, and their participation ratio, the effective number of dimensions of their covariance, of $258$--$261$ is \emph{higher} than that of every commoner band. The effect is therefore carried by the conditional-mean readout rather than by the embedding geometry. This measurement concerns the static table, not the contextual embeddings on which the flow runs.

\textbf{The read window in the decoded text.} In the text decoded at each step, the sentence frame forms at $t{\approx}0.18$, and the edits that follow are mostly to names, with a few to common words (\emph{player}$\to$\emph{manager}). In one tracked sample, one name goes \emph{Pierre}$\to$\emph{Ryan}$\to$\emph{John}, and another goes \emph{PBahna}$\to$\emph{Plenko}$\to$\emph{Bahno}$\to$\emph{Rocelli}$\to$\emph{Navelli}$\to$\emph{Specelli}$\to$\emph{Specto}. This is the name fabrication whose rate \S\ref{sec:collapse} measures, seen as it unfolds along the trajectory.

\section{Training-side probe: capability is not sufficiency}
\label{app:training}

If the missing sampling interface is the root cause, does installing one at training time restore the tail? A curriculum that teaches the model to accept given words installs the capability to use them, which we call the \emph{given-word interface}, but the capability alone does not restore the tail.

\textbf{Setup.} We trained twin 43.7M ELF-S models from scratch, with depth 8, width 512, 256-token sequences, and identical seeds and RNG streams. Each ran for one epoch of OpenWebText ($1.24$B tokens) in $75{,}812$ steps of $64$ sequences, that is, $18{,}953$ optimizer updates at gradient accumulation $4$. The \emph{base} arm is the control, and the \emph{infill} arm follows a curriculum that reveals a random subset of clean ground-truth positions as conditions, in $50\%$ of samples and with a reveal fraction drawn from $U(0,0.5)$; this is exactly the regime in which committed words are later given to the model.

\textbf{Two confounds, fixed before the reported runs.} An audit found two confounds while a first pair of runs was about a third of the way through training, at steps $24{,}600$ and $23{,}000$ of $75{,}812$. First, the reveal mask silently deleted ${\sim}12.5\%$ of the supervision of the decoding head. Because its cross-entropy training makes it spread probability over every plausible word, it is the worst head to weaken here. Second, the reveal draws consumed the shared dropout RNG and desynchronized the two arms. We patched both and retrained both arms from scratch. As expected if the fix is a no-op on the control, the patched control reproduces the unpatched run at the early logged steps (losses at steps $200$ and $400$ equal to all printed digits). All numbers below come from the patched, fully trained checkpoints.

\textbf{The capability installs, and half of it was already there.} When a fraction $f$ of the true tokens is revealed, the infill arm reconstructs the hidden positions far better than base: $24.1\%$ against $7.2\%$ at $f{=}0.1$, $50.7$ against $20.1$ at $f{=}0.25$, and $75.1$ against $44.9$ at $f{=}0.5$, with $99.4$--$99.9\%$ fidelity on the revealed positions. The base arm's conditioning input, however, already carries given words to some extent, because attention exploits given clean positions even without training. The accurate statement is therefore that the curriculum roughly \emph{doubles the efficiency} of a given-word interface that partly exists, not that it creates one from nothing: infill at $f{=}0.25$ already exceeds base at $f{=}0.5$ ($50.7\%$ against $44.9\%$).

\textbf{Capability alone does not change sampling behavior.} If the installed given-word interface let the model use a given word, committing a sampled word should raise the tail more than committing the argmax. Under mid-trajectory clamping (Appendix~\ref{app:checks}), the intervention available at the time, it does not: sampled and argmax words remain indistinguishable at every clamp fraction, the share of positions clamped, so sufficiency is falsified \emph{in that form}. The difference nonetheless has a shape. The sampled-minus-argmax tail rises with the clamp fraction in the infill arm, by $+0.33$, $+0.53$, and $+0.85$pp at fractions $0.03$, $0.08$, and $0.15$, and falls in the base arm. We read this as the given-word interface acting, with its gains eaten by the cost of replacing continuous states with discrete tokens mid-trajectory: across the same three clamp fractions, the infill arm's non-word rate rises from $3.1\%$ to $6.6\%$.

This reading motivated \graft{}'s restart form, which exercises the same given-word interface in its \emph{trained} regime, with conditions given from $t{=}0$. There, committing a sampled word does separate from committing the argmax: with four restart rounds that each commit words at the eight positions of highest decoding-head entropy, the unfiltered tail of the infill arm is higher by $0.98$ and $1.24$pp on two seeds, and that of the base arm by $0.63$pp. \graft{} itself commits the strongest losing rare candidate rather than a sample, and its argmax control is in \S\ref{sec:controls}.

\textbf{The curriculum has a side effect.} The infill arm's unconditional tail is \emph{lower} than that of base ($3.79$ against $5.23$), so infill training is not itself a tail fix. One caveat concerns scale: ELF-S is a concept-validation model trained on $1/36$ of the official training tokens, so every comparison in this appendix is made between the two ELF-S arms or within one of them, never against the official checkpoints.

\section{Negative results and retractions log}
\label{app:negative}

In the spirit of \citet{lipton2018troubling}, this appendix records the failures that shaped this work: conclusions we retracted after finding bugs of our own, interventions whose failure led to \graft{}'s restart design, variants and settings that did not help, and a first run of the discrete-line control that we excluded.

\textbf{Retracted conclusions.} Bugs of our own forced us to retract early conclusions, and two of them bear on this paper. The first was that non-word positions have a $2.8\times$ smaller gap between the decoding head's top two probabilities; it died on two bugs, a missing noise-scale factor and unmasked post-EOS garbage. The second was an apparent verbatim-memorization pathology, in which $83\%$ of ELF documents, against $75\%$ of human ones, appeared to contain a $100$-word span of training text. It was entirely an artifact of a third bug, a hash collision: a 32-bit n-gram hash at $1.2\times10^7$ entries has an $81.3\%$ per-document false-positive rate. Once the hash is widened to 64 bits, a rerun against an index of $20{,}000$ training documents (documents $2{,}000$--$21{,}999$ of the reference shard, disjoint from the human reference) \emph{reverses} the conclusion. Wherever the models match any span, human reference text matches spans of that length at $4.6$--$23\times$ their rate, because OpenWebText itself is full of boilerplate, and the models match no span of $50$ words or more. No memorization pathology therefore appears in ELF-B/M/L; LangFlow was not tested. Both retractions predate the results reported here. The EOS-masking bug did recur once, however, in a new probe where it made $51.5\%$ of the measured positions padding junk, and the same audit caught it.

\textbf{Failed interventions.} Three failed interventions shaped the method, the first two on ELF-B and the third on ELF-S, and they are the measurements behind the two cheaper alternatives that \S\ref{sec:method} rejects. The first alternative, discretizing the sequence at one intermediate step, corresponds to the all-position design of (i) below. The second, moving selected positions toward a target word over several steps, corresponds to (ii) for small moves, which are undone, and to the clamping designs of (i) for strong moves, which keep the word but multiply non-words.

\emph{(i) Mid-trajectory injection of decoded words.} Over a band of steps that contains the read window, the decoding head's word, sampled or argmax, is decoded and re-encoded into the running state, in three designs: mixing, over $t\in[0.15,0.45]$, and swapping and clamping, over $t\in[0.15,0.30]$. Mixing it into every position raises the tail, but generative perplexity explodes, from $29$ without mixing to $81$ at a mixing weight of a quarter and $699$--$2{,}057$ at a half or more. Swapping chosen positions, which replaces their state, or clamping them, which also holds them as conditions, costs less. At the lower settings of each design (a mixing weight of a quarter, swapping $5$--$15\%$ of positions, and clamping $3$--$15\%$), sampled and argmax injections are indistinguishable: clamping, for example, gives a tail of $6.5$--$9.9\%$ when the word is sampled and $7.0$--$9.3\%$ when it is the argmax, with non-words at $2.2$--$4.5\times$ the stock sampler's $2.28\%$. The two separate only at settings that have already broken the text, where sampled injection inflates both the tail and the non-words further: a mixing weight of a half gives a tail of $25.73\%$ against $14.65\%$, with non-words at $20.25\%$ against $12.10\%$. A null control that clamps the same positions without changing their words costs far less: its non-word rate is $2.48\%$ ($n{=}64$), and $3.03\%$ against the rerun stock sampler's $2.41\%$ at $n{=}200$. Most of the damage therefore comes from replacing the continuous state with re-encoded hard tokens, that is, from step-wise discretization itself. The read-time scan of Appendix~\ref{app:commitlog} shows the same failure step by step: sampled early, the positions do not compose into text, and sampled late, nothing changes.

\emph{(ii) Soft mid-trajectory pulls.} Over $t\in[0.14,0.31]$, which also contains the read window, small pulls in the style of \citet{lpir2026} move the chosen positions toward target words, with a placebo that pulls toward a randomly chosen real word. Every statistic stays flat ($4.49$--$5.00$, against the stock sampler's $4.68$), because repeated denoising, the iterated map of Appendix~\ref{app:exclusions}, absorbs transient perturbations.

\emph{(iii) Mid-trajectory clamping of a trained given-word interface.} On the infill arm of Appendix~\ref{app:training}, clamping still gives the same result for sampled and argmax words. In short, of all the designs we tried, only the restart form, with conditions given from $t{=}0$, survives, and it is the method.

\textbf{Variants that did not help.} Three variants of \graft{} were tested on LangFlow. Repeating splicing's re-decoding of the next position until no word-continuation piece remains (\emph{iterated splicing}) gives the same accept rate and accepted-set tail as the single pass of \S\ref{sec:method}, with identical outputs for $178$ of $200$ samples, so one pass suffices. Never committing the same word type twice in a sample (\emph{dedup}) halves the commitments ($2{,}437\to1{,}202$) and cuts the gain from $+2.66$ to $+0.80$pp, so the commitment count is the dose. Leaving the self-conditioning input unfixed at committed positions (\emph{scraw}) helps alone ($+3.87$) but interferes with co-pinning, giving $+2.92$ in combination against co-pinning's $+5.33$. We read this as interference between two changes that both act on the self-conditioning input at the fixed positions. A fourth change fails on ELF: the step-schedule shift used on ELF-B does not transfer to the larger models. On the same seed, it lowers accept on both (ELF-M $67.0\to64.5\%$, ELF-L $54.0\to51.0\%$), leaves the ELF-M tail unchanged ($11.54\to11.61$), and lowers the ELF-L tail ($12.62\to10.69$), which is why the ELF-M and ELF-L champions omit it.

\textbf{Overdosing the anti-repetition correction.} Too large a dose of the correction breaks both models off-manifold, with different symptoms, and this breakdown bounds the usable range of $\lambda$. On ELF at $\lambda{=}2$, non-words rise from $2.35$ to $4.4$--$4.5$. On LangFlow at $\lambda{=}10$, 4-gram repetition explodes to $71.5$ and types collapse to $1{,}743$ per 100k, while non-words fall.

\textbf{An excluded LLaDA run.} A first-pass LLaDA control, with 16-token prefixes, one 512-token block in 256 steps, low-confidence remasking, and temperatures $0$, $0.5$, and $1$, degenerated into repetition (median 4-gram repetition of $70$--$85\%$), and we exclude it rather than report it. Short smoke tests (128 tokens, 4-gram repetition $4.3$) had masked the repetition, which accumulates over full 512-token generations. The rerun of Table~\ref{tab:appllada}, with semi-autoregressive blocks of $64$ and $512$ steps, supersedes it and locates the cause in the remasking rule rather than in the model: \emph{low-confidence} remasking is itself semi-greedy and stays repetitive in blocks too (4-gram repetition $64$--$81\%$), while \emph{random-order} remasking, the fair ``interface fully open'' arm, shows no repetition.

\section{Additional results}
\label{app:results}

This appendix collects the supporting results, roughly in the order of the main text. Appendices~\ref{app:buildup}--\ref{app:commitlog} detail \graft{}'s components and cost, the reference arms, and the commitments; Appendices~\ref{app:heads}--\ref{app:qdecile} support the diagnosis, the mechanism, and the comparisons of \S\ref{sec:collapse}, \S\ref{sec:cause}, and \S\ref{sec:setup}; Appendices~\ref{app:foreign}--\ref{app:buckets} check the metrics, the gates, and the dose; and Appendix~\ref{app:repro} collects what is needed to reproduce Table~\ref{tab:main}.

\subsection{Component build-up, cost, and the anti-repetition correction}
\label{app:buildup}

This subsection breaks \graft{} into its components on ELF-B, gives its cost, and reports the anti-repetition correction and the ladder controller.

\textbf{Component build-up on ELF-B.} Each step below gives the unfiltered tail and non-word rate (\%) behind Table~\ref{tab:ablation}, with the fidelity in parentheses. The stock sampler reads $4.54/2.35$. Rare commitment with no rule reads $7.92/4.14$ ($86.7\%$), adding atomic commitment gives $8.86/2.71$ ($87.4\%$), and adding the \emph{write-back}, which copies each committed token into the output, gives $9.20/2.74$ ($100\%$). Splicing as defined in \S\ref{sec:method}, which adds the re-decoding of the next position to this write-back, gives $9.61/2.47$ ($100\%$), and word-boundary co-pinning then gives $9.10/1.78$ ($100\%$). None of these steps uses the anti-repetition correction or the gates, and the last one also uses the step-schedule shift. The argmax controls of the first two steps read $4.99/3.30$ and $4.39/2.50$. Against the matching argmax control, the conversion rate, the tail gained per non-word added, is $3.5{:}1$ for commitment through restarts on ELF-B and $21{:}1$ with atomic commitment. On the ELF-S infill arm, it is $0.36{:}1$ for mid-trajectory clamping and $4.3{:}1$ for commitment through restarts (Appendix~\ref{app:training}).

\textbf{Under the acceptance bars.} Under the accept-then-measure protocol with the single bar, read as accept rate and \tailacc{}, the stock sampler reads $69.0\%$ and $4.31$. An arm that commits sampled words at all $1{,}024$ positions, rather than at eight per round, reads $14.5\%$ and $11.50$: a human-level tail at an unusable accept rate, which marks the far end of the trade-off. Atomic commitment reads $62.0\%$ and $8.59$ ($+4.27$), atomic commitment with the write-back $60.0\%$ and $8.81$ ($+4.50$), and the argmax control $67.5\%$ and $4.14$ ($-0.18$, $p{=}0.50$). Under the dual bar, the gain of atomic commitment with the write-back grows with $m$, reaching $+1.13$, $+2.44$, and $+4.42$ at $m{=}2$, $4$, and $8$. Over the write-back alone, splicing raises accept from $45.5$ to $51.0\%$ and the tail from $8.72$ to $9.24$.

\textbf{Cost.} \graft{} needs no training, and its sampling cost can be counted in two ways. Counted in denoising steps, an arm runs four partial trajectories up to the read time and one full trajectory. Because ELF's logit-normal schedule places many steps early, the read time falls $34.8\%$ of the way through the steps on ELF-B (with the step-schedule shift) and $59.5\%$ on ELF-M/L. An arm therefore costs $2.4$ full trajectories on ELF-B, $3.4$ on ELF-M/L, and $2.1$ on LangFlow's linear 256-step grid. The figure of between two and three trajectories quoted in \S\ref{sec:method} thus holds on ELF-B and on LangFlow, while ELF-M/L need $3.4$, still fewer than the four samples of the equal-budget baseline. Measured as wall clock, an earlier ELF-B configuration (the champion, with the anti-repetition correction and the step-schedule shift but no co-pinning) took $59$\,s against the stock sampler's $27$\,s for $n{=}64$ samples, or $2.2\times$. These whole-process times include model loading, which understates the ratio.

\textbf{The price in clean-PPL.} The coherence price, measured as clean-PPL, comes mostly from the anti-repetition correction rather than from commitment. Commitment alone at $m{=}8$ costs $+7.3$ over ELF-B's $26.47$, whereas the champion, which adds the correction and the step-schedule shift, reaches $45.6$--$50.4$ across three seeds. Champion$+$co-pinning \emph{without} the correction is thus a cheaper operating point, which keeps accept above stock ($67.5\%$ against $66.0\%$) and most of the tail gain ($+4.61$ of $+5.39$). Part of this price is the metric's own bias, since on LangFlow the argmax control's clean-PPL \emph{drops} ($47.72\to44.90$); not all of it is, since human text carries an $11.2\%$ tail at a clean-PPL of $16.66$. Across components, the gains decompose along the conversion rate $\Delta$tail\,:\,$\Delta$non-word that ruled out mid-trajectory clamping (above), and three plausible variants fail informatively (Appendix~\ref{app:negative}): iterated splicing gives the same readings as splicing, \emph{dedup} halves commitments and cuts the gain by more than two thirds, and \emph{scraw} interferes with co-pinning.

\textbf{The anti-repetition correction.} The repetition direction $\mathbf{u}$ of \citet{lpir2026} is stable: independent estimates agree at cosine $0.948$, and the projection on $\mathbf{u}$ correlates with repetition at $0.469$--$0.506$ on ELF and $0.473$ on LangFlow. It transfers from ELF-B to ELF-M and ELF-L unchanged. On ELF, $\lambda$ values of $0.25$, $0.35$, and $0.5$ form a plateau (accept $51.5$--$53.5\%$), while $\lambda{=}2$ lies beyond the usable range (accept $19.5$--$25\%$, non-words $4.4$--$4.5\%$). On LangFlow, the sweet point is $\lambda{=}0.5$ (accept $39.5\to43.5\%$, tail $10.94$); repetition falls monotonically up to $\lambda{=}3$, and $\lambda{=}10$ reverses catastrophically.

\textbf{The ladder.} The ladder tries its levels from strongest to weakest for each sample and keeps the first output that passes both bars, or else the stock output, so it costs at most one generation per level. The ELF ladder has four levels---champion$+$co-pinning, the champion, $m{=}4$ with splicing and the correction, and stock---which are used $139$, $18$, $10$, and $33$ times out of $200$ on the first seed. Across seeds, accept is $88.5$, $89.0$, and $88.0\%$, and \tailacc{} is $9.48$, $9.04$, and $9.16$; their means form the ladder row of Table~\ref{tab:appaccept}. The LangFlow ladder has five levels---champion$+$co-pinning, the champion, champion$+$co-pinning with the self-conditioning input left unfixed, champion$+$co-pinning at $m{=}4$, and stock---which are used $104$, $20$, $7$, $2$, and $67$ times.

\subsection{Reference arms and unfiltered rates}
\label{app:refarms}

This subsection collects two reference tables. Table~\ref{tab:apparms} gives the autoregressive arms behind the contrast of \S\ref{sec:collapse}, showing what a decoding rule alone reaches on GPT-2 under the full accept-then-measure protocol with the dual bar. Table~\ref{tab:appunfiltered} gives the unfiltered tail and non-word rates of the earlier, ungated configurations, together with their top-5k tail (Appendix~\ref{app:checks}).

\begin{table}[ht]
\centering
\small
\caption{Autoregressive reference arms under the full accept-then-measure protocol with the dual bar ($n{=}496$--$500$; greedy, $n{=}100$). Pure sampling reaches the deepest tail at the worst quality, and nucleus sampling is the quality point. The contrast of \S\ref{sec:collapse} concerns whether such a knob exists, not the quality at its extremes.}
\label{tab:apparms}
\begin{tabular}{lrrrr}
\toprule
Arm & accept\,\% & \tailacc{} & tail (unfiltered) & non-word (unfiltered) \\
\midrule
GPT-2 greedy            & 0.0  & ---   & 0.00  & 0.00 \\
GPT-2 top-$k$ 50        & 90.3 & 5.28  & 5.56  & 0.24 \\
GPT-2 nucleus $0.95$    & 84.9 & 9.80  & 10.06 & 1.17 \\
GPT-2 sample $T{=}1$    & 51.7 & 13.36 & 14.60 & 3.01 \\
GPT-2 medium greedy        & 0.0  & ---   & 0.26  & 0.00 \\
GPT-2 medium top-$k$ 50    & 92.6 & 5.92  & 6.14  & 0.25 \\
GPT-2 medium nucleus $0.95$& 85.2 & 11.41 & 11.87 & 1.23 \\
GPT-2 medium sample $T{=}1$& 40.4 & 15.35 & 16.28 & 3.73 \\
\midrule
Human (OpenWebText reference)   & 90.5 & 11.17 & 11.49 & 0.66 \\
\bottomrule
\end{tabular}
\end{table}

\begin{table}[ht]
\centering
\small
\caption{Left: unfiltered tail and non-word rates, with no acceptance filter. Right: the top-5k tail of Appendix~\ref{app:checks} on the accepted set (dual bar). The tail gains survive without filtering, and unfiltered non-words do not rise with co-pinning or the ladder, although they do for the two champion rows. Orderings are unchanged under the top-5k tail, on which all three non-stock LangFlow rows overshoot the human level of $12.35$; we report the overshoot as measured.}
\label{tab:appunfiltered}
\begin{tabular}{lrr|r}
\toprule
Arm & tail (unfiltered) & non-word (unfiltered) & top-5k tail \\
\midrule
LangFlow stock          & 8.59  & 2.44 & 11.05 \\
LangFlow champion       & 15.09 & 3.27 & 13.70 \\
LangFlow champion$+$co-pinning & 16.20 & 2.41 & 16.48 \\
LangFlow ladder (5)     & 12.35 & 1.98 & 15.87 \\
\midrule
ELF-B stock             & 4.54  & 2.35 & 6.61 \\
ELF-B champion          & 9.45  & 2.52 & 11.26 \\
ELF-B champion$+$co-pinning & 10.03 & 2.22 & 11.62 \\
ELF-B ladder (4)        & 9.04  & 2.00 & 11.42 \\
\midrule
Human                   & 11.49 & 0.66 & 12.35$^{\ast}$ \\
\bottomrule
\multicolumn{4}{l}{\footnotesize $^{\ast}$human top-5k tail on the accepted set (accept $90.5\%$).}
\end{tabular}
\end{table}

\subsection{What the commitments look like, and where the read window sits}
\label{app:commitlog}

This subsection describes the logged commitments, how the read time was chosen and where it sits in each model, and where committing costs non-words. It closes with the properties that ELF and LangFlow share.

\textbf{The commitment logs.} Every commitment records its round, position, token, and counterfactual, both probabilities, the entropy, the Zipf frequencies of both tokens, the final token, and whether it was kept. On ELF-B, the ungated champion$+$co-pinning arm makes $6{,}220$ commitments (${\sim}31$ per sample) and $6{,}220$ co-pinned positions, against $6{,}132$ commitments for the gated \graft{} arm of Table~\ref{tab:main} on the same seed. Splicing enforces $100\%$ fidelity. The committed tokens are $100\%$ rare and $0\%$ non-words, with a mean entropy of $3.35$ nats, whereas the counterfactual is never a rare word ($0$ of $6{,}220$) and is a non-word at $20$ of those positions. The median 4-gram repetition is $1.48$, against $2.49$ for the stock sampler. On LangFlow, the ungated champion$+$co-pinning arm makes $2{,}429$ commitments (${\sim}12$ per sample), against $2{,}403$ for the gated arm on the same seed, again with $100\%$ fidelity, all committed tokens rare, and a mean entropy of $3.29$.

In the earlier ELF-S runs, fidelity on the infill arm rises with round, from $77$ to $86\%$, as later commitments embed in more settled drafts, while the base arm stays at $62$--$67\%$. These runs also show why \graft{} selects positions as it does: sampling at the positions of highest decoding-head entropy is nearly blind there, with a median committed-token probability of $0.007$. \graft{} therefore selects the positions where a rare candidate is losing (Step~3 of \S\ref{sec:method}) rather than selecting by entropy. Excerpts with the committed words marked are in Appendix~\ref{app:qualitative}.

\textbf{Choosing the read time.} The read time is fixed before any reported number is measured, by a scan that asks two questions at every step of the trajectory: how much is left to sample, and whether what would be sampled is still coherent. At each of the $32$ denoising steps (batch $16$), we read the decoding head's per-position distribution and record its entropy and top-1 probability; separately, we sample every position independently at that step and score the resulting text for tail and non-word rate. The two ends of the trajectory fail for opposite reasons. Early on, the tail is still decidable, but the per-position distributions are broad, so sampling all positions independently factorizes the joint distribution and the text falls apart. Late in the trajectory, the distributions are sharp and coherent, but the words are already fixed, so sampling changes nothing; on ELF, the frequency profile is frozen after $t{\approx}0.25$. The read window is the interval in which both conditions hold, under each model's own time convention. ELF, whose time runs from noise at $t{=}0$ to data at $t{=}1$, settles content at $t{\approx}0.17$--$0.25$ and is read at $t_{\mathrm{read}}{=}0.20$. LangFlow's time runs the other way: its per-position distributions stay broad through $t{\approx}0.85$--$0.75$ and have largely settled by $t{\approx}0.55$, so it is read at $t_{\mathrm{read}}{=}0.72$, after the broad phase, where the draft has formed but sampling room remains.

Curves of when the decoding head's entropy is used up on ELF, resolved by word frequency and computed on noised human text rather than on the model's own trajectories, place common-word decisions at $t{\approx}0.20$--$0.31$ and rare-word decisions later and over a wider span ($t{\approx}0.25$--$0.40$). This does not contradict the frozen profile above, which describes the model's own trajectories, already short of rare words by $t{\approx}0.25$. ELF's default logit-normal schedule, however, puts its median step at $t{\approx}0.18$ ($t{\approx}0.27$ with the step-schedule shift used on ELF-B), so step density is misaligned with exactly the decisions that need it; this is the mechanistic reading behind the step-schedule shift.

\textbf{Where the non-word cost sits.} Committing words costs non-words on both models, but the cost sits in different places. On ELF, it is local. With distance $\delta$ measured in word slots from the nearest committed position ($\delta{=}0$ is the committed slot itself), word-level non-words spike at committed slots ($36.07\%$ at $\delta{=}0$ before atomic commitment and $7.88\%$ after) and stay at the background rate in the far field. The cause is sub-word gluing, in which a committed token fuses with a neighboring piece into a non-word. On LangFlow, by contrast, the cost is global. Committed slots are perfectly clean ($0.00\%$ at $\delta{=}0$), while the rate rises uniformly over the rest of the text, to $4.7$--$5.4\%$. Over whole texts, with token ids re-derived from each arm's text, the same word-level rate is $4.50$ for stock, $4.92$ for the argmax control, and $5.19$ for the champion. Roughly $60\%$ of LangFlow's extra non-words therefore come from fixing words and restarting as such, a cost that the argmax control pays too; about $40\%$ come from the rare content of the commitments, and none from gluing at the committed slots.

\textbf{Nine properties shared by ELF and LangFlow.} In sum, nine properties agree between the two models. Both show the tail collapse, with the conditional-mean readout as its mechanism and committed words as its repair. In both, committing a sampled word separates from committing the top choice, and fabricated proper names are the collapse's signature. Committing costs non-words on both, as local sub-word gluing on ELF and as a global rise on LangFlow (above). Both show commitment-induced repetition and the repetition direction of \citet{lpir2026}, and in both that correction has a finite usable range of $\lambda$, closed by an off-manifold breakdown.

\subsection{The two heads, scale, and the baseline comparisons}
\label{app:heads}

This subsection collects analyses that support claims across \S\ref{sec:cause} and \S\ref{sec:setup}: the two-head comparison, model scale, clean-PPL on every model, the two nearest samplers, best-of-$K$ curves under a strict budget, the v2 judge, and two conditional settings, prefix continuation and the LLaDA control.

\begin{table}[ht]
\centering
\small
\caption{Per-position error rate of the denoising head divided by that of the decoding head, per Zipf band, on the same network and the same input at noise level $t{=}0.5$ on ELF's time axis (matched input, with the decoding head's time argument at $0.714$). The last row gives the ELF-B error rates behind the ratios (denoising / decoding, \%), which Figure~\ref{fig:heads} shows. The setup and the other three noise levels are given in the text.}
\label{tab:apptwoheads}
\begin{tabular}{lccccc}
\toprule
Excess error (denoising / decoding) & Zipf $\geq 6$ & $[4,6)$ & $[3,4)$ & $[2,3)$ & $<2$ \\
\midrule
ELF-B & $1.28\times$ & $1.91\times$ & $2.99\times$ & $2.83\times$ & $2.93\times$ \\
ELF-M & $1.00\times$ & $1.23\times$ & $1.91\times$ & $2.20\times$ & $2.25\times$ \\
ELF-L & $1.45\times$ & $1.47\times$ & $2.21\times$ & $2.51\times$ & $1.97\times$ \\
\midrule
ELF-B error rates (\%) & $1.0/0.7$ & $9.3/4.9$ & $21.9/7.3$ & $19.8/7.0$ & $16.1/5.5$ \\
\bottomrule
\end{tabular}
\end{table}

\textbf{Setup and noise level of the two-head comparison.} The two heads are compared at the matched input, where their inputs are equal up to a positive scalar (Appendix~\ref{app:exclusions}). Each scale reads the first $96$ reference documents, about $75{,}000$ word positions, of which $547$ fall in the rarest band. Section~\ref{sec:cause} reads the comparison at $t{=}0.5$ (Table~\ref{tab:apptwoheads}), the latest of the four noise levels we measured ($t{=}0.25$, $0.31$, $0.4$, and $0.5$). At that level, the excess is largest in the three rare bands at every scale ($1.9$--$3.0\times$, against $1.0$--$1.9\times$ in the two common bands) and largest on ELF-B. The excess builds up along the trajectory. At $t{=}0.4$, it is already present in the rare bands but smaller, with $[3,4)$ ratios of $1.64\times$ on ELF-B, $1.14\times$ on ELF-M, and $1.22\times$ on ELF-L. At $t{\leq}0.31$, both heads still miss most rare words, with error rates above $60\%$ in every rare band, and there is no excess: the ratios are at most $1.14$ on ELF-B in any band and below $1$ in every band on ELF-M and ELF-L, where the decoding head is the less accurate of the two. The rare-word excess of \S\ref{sec:cause} is therefore a property of the later part of the trajectory, and \graft{}'s read time, $t{=}0.20$, lies before it.

\begin{table}[ht]
\centering
\small
\caption{Gains across model scale under the accept-then-measure protocol with the dual bar (ungated runs; the champion runs without the step-schedule shift on ELF-M and ELF-L). The champion column gives the range over three seeds, and the champion$+$co-pinning columns give the first seed. ELF-M and ELF-L reach and pass the human accepted-set tail of $11.17$. The ELF-S row is an unfiltered tail gain rather than an accepted-set gain (see text).}
\label{tab:appscale}
\begin{tabular}{lrrrr}
\toprule
Model & champion $\Delta$\tailacc{} & \multicolumn{3}{c}{champion$+$co-pinning:\ \ accept\,\% / \tailacc{}\,\% / $\Delta$} \\
\midrule
ELF-S (43.7M) & $+0.9$ (unfiltered, ${\sim}1$/sample) & \multicolumn{3}{c}{---} \\
ELF-B (105M)  & $+4.59$ to $+4.92$ & 69.5 & 9.68 & $+5.39$ \\
ELF-M (342M)  & $+5.96$ to $+6.30$ & 64.0 & \textbf{11.81} & $+6.58$ \\
ELF-L (652M)  & $+4.86$ to $+5.95$ & 63.5 & \textbf{12.98} & $+6.32$ \\
\bottomrule
\end{tabular}
\end{table}

\textbf{Scale.} Table~\ref{tab:appscale} gives the gains at every scale. Over three seeds, the champion$+$co-pinning gains are $+6.58$, $+6.25$, and $+6.41$ on ELF-M, and $+6.32$, $+5.51$, and $+5.72$ on ELF-L. Gains are larger on ELF-M than on ELF-B under both configurations, while ELF-L lies between the two on average for the champion and above ELF-B with co-pinning. ELF-S ($43.7$M parameters, $1.24$B training tokens) brackets the mechanism from below. Its decoding head is weaker: the AUC with which $q$ predicts word loss (Appendix~\ref{app:qdecile}) is $0.82$--$0.83$ over all positions and $0.69$--$0.73$ on rare positions, against $0.88$ and $0.84$ on ELF-B. It finds about one qualifying position per sample, and the gain collapses accordingly. Its row reports the unfiltered tail gain of atomic commitment with splicing ($3.79\to4.70\%$ on the infill arm), not an accepted-set gain.

\textbf{Clean-PPL on every model.} Clean-PPL, against a human value of $16.66$, reads as follows. On ELF-B, stock reads $26.47$, the $m{=}2$, $4$, and $8$ arms read $27.22$, $28.41$, and $33.77$, and the champion reads $45.64$--$50.41$ over three seeds; champion$+$co-pinning reaches $46.83$ against the same scoring batch's stock $26.47$. On ELF-M, stock and the champion read $39.11$ and $58.14$, and on ELF-L, $39.36$ and $60.08$. On LangFlow, stock reads $47.72$, the champion without the anti-repetition correction $50.71$, the argmax control $44.90$, champion$+$co-pinning $55.86$, and the ladder $54.21$.

\begin{table}[ht]
\centering
\small
\caption{Comparison with MCB on the same LangFlow checkpoint, under our protocol. Stock and MCB run at an equal number of function evaluations (NFE), and MCB is our port of the authors' released sampler to this checkpoint. The champion$+$co-pinning row (ungated) costs about $2.1$ full trajectories (Appendix~\ref{app:buildup}) and is its seed-$42$ run, the seed of the MCB arms; over three seeds, its gain is $+3.07$ to $+5.33$, and the gated \graft{} arm gains $+4.80$ on seed $42$ (Table~\ref{tab:applfdose}). In the Sampler column, $p$ is MCB's top-$p$ threshold; in the $\Delta$ column, it is a bootstrap $p$-value. At every temperature we could run, MCB trades cleanliness for tail in the wrong direction.}
\label{tab:appmcb}
\setlength{\tabcolsep}{4.5pt}
\begin{tabular}{lrrlrr}
\toprule
Sampler & accept\,\% & \tailacc{} & $\Delta$\tailacc{} [95\%~CI] & \shortstack{4-gram\\rep.} & non-word\,\% \\
\midrule
Stock                & 60.5 & 8.28 & --- & 0.53 & 2.44 \\
MCB $T{=}1.0$        & \textbf{77.9} & 7.77 & $-0.51$ [$-1.02,-0.02$], $p{=}.038$ & \textbf{0.00} & 1.87 \\
MCB $T{=}0.8$, $p{=}0.9$ & 19.8 & 4.77 & $-3.51$ [$-4.49,-2.48$] & 20.10 & 0.74 \\
MCB $T{=}0.6$, $p{=}0.9$ & ${\sim}0$ & --- & off-manifold breakdown & 35.26 & 4.23 \\
champion$+$co-pinning & 52.0 & \textbf{13.61} & $\mathbf{+5.33}$ [$+3.40,+7.46$] & 1.28 & 2.41 \\
\bottomrule
\end{tabular}
\end{table}

\textbf{The nearest samplers.} MCB and INCO are the two samplers closest to \graft{} that were proposed for continuous-space language models. Table~\ref{tab:appmcb} compares MCB, which samples an endpoint at every step and bridges toward it, with the ungated champion$+$co-pinning arm. INCO \citep{inco2026} instead redistributes denoising progress across positions by inverse confidence and never draws a token. We ran it on the same LangFlow checkpoint at equal NFE ($256$ steps, seed $42$, $n{=}200$), at a confidence temperature $\tau_{\mathrm{conf}}$ of $1$, $2$ (the authors' setting), and $4$. Under the protocol of Table~\ref{tab:appmcb}, against a stock run of the same seed on the same machine, its accepted-set tail changes by $-0.04$, $-0.00$, and $-0.07$ ($p{=}0.94$, $0.97$, and $0.85$), at accept rates of $58$--$60\%$. Its Zipf profile is unchanged as well, with a JSD to human of $.0025$ against stock's $.0024$. As a check on our port, its setting with equal weights for all positions reproduces the stock sampler token for token on eight samples. MCB resamples at every step and INCO reallocates steps, and neither raises the tail; on the same seed, committing from the first step raises it by $+5.33$ (champion$+$co-pinning) and $+4.80$ (gated \graft{}).

\textbf{Best-of-$K$ curves under a strict budget of passes.} Every best-of-$K$ selection in this paper, including the stock best-of-4 row of Table~\ref{tab:main}, uses one rule: among the $K$ candidates for a sample, it keeps the one with the highest per-sample tail share among those that pass both bars, or, if none passes, the one with the fewest non-words. The rule selects on the tail itself, which favors the baseline. We count cost in passes, where a pass is one trajectory, full or partial, so one arm of a \graft{} variant counts as $5$ passes and a stock sample as $1$. This overstates the arm's cost, which amounts to $2.1$--$3.4$ full trajectories (Appendix~\ref{app:buildup}). The comparison pools are built from the ungated arms of the ladder (Appendix~\ref{app:buildup}). On ELF-B, they add, in order, champion$+$co-pinning, the champion, $m{=}4$ with splicing and the correction, and the stock output; on LangFlow, they add champion$+$co-pinning, the champion, champion$+$co-pinning with the self-conditioning input left unfixed, and the stock output. The stock output is a candidate under the same rule, not a fallback.

We report each reading as (accept\,\%, \tailacc{}). On ELF-B, stock best-of-$K$ at $K{=}1$, $2$, $4$, $8$, and $16$ reads $(66.0, 4.30)$, $(88.0, 4.86)$, $(98.0, 5.64)$, $(100, 6.70)$, and $(100, 7.75)$, while pools of $1$ to $4$ members, costing $5$, $10$, $15$, and $16$ passes, read $(69.5, 9.68)$, $(78.5, 10.15)$, $(83.5, 10.17)$, and $(88.5, 9.89)$. On LangFlow, stock reads $(60.5, 8.28)$, $(87.5, 8.70)$, $(99.0, 9.73)$, $(100, 10.97)$, and $(100, 12.23)$, and the pools read $(52.0, 13.61)$, $(62.0, 14.41)$, $(65.5, 14.82)$, and $(84.0, 13.61)$. Each doubling of $K$ adds $0.6$--$1.1$pp of stock tail on ELF-B and $0.4$--$1.3$pp on LangFlow, and no stock reading in this range reaches the smallest pool. At the 16-pass budget, a bootstrap that resamples sample indices jointly for the two arms puts the difference between the full pool and stock best-of-16 at $+2.14$ [$+1.67$, $+2.59$] on ELF-B but at $+1.38$ [$-0.10$, $+3.02$] ($p{=}0.068$) on LangFlow. At this budget, the stock sampler thus becomes statistically indistinguishable from the pool on LangFlow, but not on ELF-B. Only the 16-pass budget was tested in this way, and the pool consists of ungated \graft{} variants rather than the gated \graft{} of Table~\ref{tab:main}.

On the unfiltered, all-token reading of Table~\ref{tab:main}, LangFlow's stock best-of-4 reaches a $10.21\%$ tail ($0.86\times$ human) with $2.43\%$ non-words. On ELF-B, the same reading gives stock best-of-4 a $5.97\%$ tail ($0.50\times$), \graft{} at the fixed dose $9.76\%$ ($0.82\times$), and \graft{} at the tuned dose $11.82\%$ ($1.00\times$; seeds $11.88$, $11.82$, and $11.76$; non-words $2.59\%$). Section~\ref{sec:controls} compares best-of-4 with the fixed dose on ELF-B and with the tuned dose on LangFlow.

\textbf{The appropriateness judge in the draft context.} Re-scored in the selection-time draft context (Appendix~\ref{app:checks}), the judge reads as follows. On ELF-B, $6{,}164$ commitments have a median of $-11.79$ against $-12.37$ for human rare words, and $57.2\%$ of them lie above the human median (v1: $52.9\%$). On LangFlow, $2{,}420$ commitments have a median of $-7.27$ against $-12.38$, and $80.6\%$ lie above it (v1: $49.6\%$). Both runs score the same $9{,}933$ human positions, each in its own scoring run, and the two human medians differ in the second decimal place. The LangFlow v2 figure partly reflects rarity composition: committed words cluster at the lighter end of rare, each backed by $q_{\mathrm{rare}}{>}0.08$, while human rare positions include heavy-tail vocabulary. This is why we report v1 and v2 side by side rather than v2 alone. In both models, draft-context counterfactuals score a higher raw log-probability than the commitments, as the frequency prior dictates, so the calibration against human rare words is the like-for-like readout.

\textbf{Prefix-conditioned generation.} In this setting, each model continues a 16-token human prefix taken from an OpenWebText document: ELF and LangFlow continue to their $1{,}024$-token length and LLaDA to $512$ tokens, with $n{=}200$ on ELF and LangFlow and $100$ on LLaDA. On ELF, the prefix-conditioned stock sampler is healthy (unfiltered tail $4.80$, non-words $2.17$), and champion$+$co-pinning wins on both axes (accept $71.5\%$, \tailacc{} $9.81$, $\Delta{+}5.28$ [$+4.83$, $+5.71$]). LangFlow has no native conditioning input, so the prefix is simulated by holding all $16$ contiguous tokens fixed. The text then floods with prefix echoes, with a median 4-gram repetition of $44.1$, and leaving the self-conditioning input unfixed does not rescue it ($41.5$). The echo therefore does not travel mainly through the self-conditioning input; it appears to be carried by attention-level coordination around the fixed span. At \graft{}'s scattered dose of eight commitments per round, no such echo appears, so $16$ \emph{contiguous} fixed positions form a regime of their own.

\textbf{The LLaDA control.} Table~\ref{tab:appllada} gives the LLaDA control behind the LLaDA range of Figure~\ref{fig:fig1}b, in which the model's native sampling interface is either shut or opened.

\begin{table}[ht]
\centering
\small
\caption{LLaDA-8B discrete-line control (semi-autoregressive decoding in blocks of 64, 16-token human prefixes, 512 tokens, $n{=}100$). The comparison is within the model: LLaDA is not trained on OpenWebText, so its bars are cross-domain. Opening the native sampling interface decides the outcome: random-order remasking at $T{=}1$ reaches the human tail with zero repetition, while the semi-greedy schedules do not. Types per 100k fall with sample size, so in that column the human row ($1{,}000$ documents) is not comparable to the LLaDA rows ($100$ samples). The \emph{argmax} arm decodes at temperature $0$, which shuts the sampling interface; the other rows name the remasking rule of Appendix~\ref{app:negative}, low-confidence or random-order, and its temperature; and n.s.\ marks a $\Delta$ whose $95\%$ interval includes zero.}
\label{tab:appllada}
\setlength{\tabcolsep}{4.5pt}
\begin{tabular}{lrrrrl}
\toprule
Arm & \shortstack{tail\\(unfiltered)} & \shortstack{4-gram\\rep.} & types/100k & \shortstack{accept\\(\%)} & $\Delta$\tailacc{} vs.\ argmax \\
\midrule
argmax (interface shut)      & 6.89  & 80.9 & 5{,}755  & 7.2  & (reference) \\
low-confidence, $T{=}0.5$ & 6.85  & 77.6 & 6{,}002  & 13.5 & $-1.56$, n.s. \\
low-confidence, $T{=}1$   & 7.15  & 63.8 & 7{,}919  & 17.0 & $+0.46$, n.s. \\
random-order, $T{=}1$ & \textbf{11.16} & \textbf{0.0} & 28{,}599 & 55.8 & $+3.41$ $[+1.01,+5.96]$, $p{=}0.002$ \\
\midrule
Human (OpenWebText)                 & 11.49 & ---  & 5{,}497  & 90.5 & \\
\bottomrule
\end{tabular}
\end{table}

\subsection{Downstream: entity fabrication in summarization}
\label{app:xsum}

Committing rare words could in principle worsen the fabrication that the diagnosis started from (\S\ref{sec:collapse}). We therefore ran rare commitment with co-pinning, that is, champion$+$co-pinning without the anti-repetition correction and without the step-schedule shift, on the XSum-conditioned ELF-B checkpoint and scored entities against the source article with a per-document paired bootstrap ($n{=}200$, the same article for each arm).

\textbf{Pipeline.} These arms, the stock arm included, run on our probe pipeline, which takes $32$ ODE steps with classifier-free guidance on the source fixed at $1$, that is, switched off. ELF's own XSum sampler, used for the fabrication counts of Appendix~\ref{app:tasks}, takes $64$ ODE steps at guidance $2$. The two pipelines produce very different summaries: on the same $200$ articles, the stock arm's unsupported-entity share is $64.1\%$ here, against $28.55\%$ for ELF's sampler ($25.90\%$ over all $500$ articles it was run on). Because the two differ in step count, in guidance, and in the weights they load (Appendix~\ref{app:configs}), these runs cannot say how much of the gap comes from each, and the readings below compare arms within the probe pipeline only.

\textbf{Result.} Unsupported entities per entity change by $-2.71$pp [$-6.48$, $+0.94$], unsupported entities per $100$ words by $+0.45$ [$-0.61$, $+1.53$], and the share of summaries containing at least one by $-4.50$pp [$-9.00$, $0.00$]. All three changes are null. Their opposite signs are a length effect: summaries shorten from $16.0$ to $14.1$ words, so a per-entity drop and a per-100-word rise are the same fact.

\textbf{Dose.} This null must be read together with the dose at which it was measured. Besides lacking the correction and the step-schedule shift, the arm runs without the gates, and on short, source-conditioned summaries the read time offers almost no qualifying positions: at the default $q_{\min}{=}0.08$, the arm commits a single word across all $200$ summaries. Of its summaries, $66$ are identical to the stock sampler's, and the rest differ because each restart re-encodes the source article together with the latest draft, which changes the conditioning even without new commitments (Appendix~\ref{app:procedure}). Relaxing the selection raises the dose only modestly, to $8$ commitments at $q_{\min}{=}0.01$ and to $80$ at $q_{\min}{=}0.01$ with the read time moved to $t{=}0.35$. At $80$ commitments, the unsupported share per entity is $63.9\%$, against the stock sampler's $64.1\%$; these are point estimates, since no bootstrap was run for the relaxed arms. At the default selection threshold, rare commitment thus leaves summarization essentially unchanged, in either direction. This experiment, however, does not test the full \graft{} configuration, which adds the correction, the gates, and on ELF-B the step-schedule shift, nor whether committing rare words at the dose of unconditional generation affects entity fabrication.

\subsection{The sampler sweep: what the knobs reach, and what generative perplexity reads}
\label{app:sweep}

This subsection gives the measurements behind the reachability claim and the reading of generative perplexity in \S\ref{sec:collapse}, and behind the effective-candidate count of \S\ref{sec:intro}. The sweep covers three groups of ELF sampler settings. On ELF-B, five step-and-guidance configurations were run with $300$ samples each: 16, 32, and 64 steps at self-conditioning guidance $g{=}3$, and $g{=}1$ and $5$ at 32 steps. All use SDE churn $\gamma{=}1.5$, except the 64-step run, which uses $\gamma{=}1.0$. A separate ELF-B batch compares argmax with sampling at the final step, at temperatures up to $2.0$. On ELF-M and ELF-L, ELF's second official sampler preset (64 steps, SDE churn $\gamma{=}1.0$) was run with $1{,}000$ samples each, whereas the main experiments use the first preset (32 steps, $\gamma{=}1.5$). Table~\ref{tab:buckets} takes its ELF-B column from a 32-step batch at the first preset and its ELF-M and ELF-L columns from these 64-step batches, so its scales differ in sampler preset as well as in size; at the first preset, ELF-L reads $7.42\%$ on the probe pipeline (below). Tail shares in this subsection follow the alphabetic rule and are read against the human rate of $11.49\%$ under that rule, as in Table~\ref{tab:buckets}. Table~\ref{tab:main} instead uses the all-token rule and a human rate of $11.88\%$, so ratios from the two conventions are close but not interchangeable (Appendix~\ref{app:metrics}).

\begin{table}[ht]
\centering
\small
\renewcommand{\arraystretch}{0.92}
\caption{The sampler sweep on ELF, under the alphabetic rule. The ELF-B step-and-guidance rows have $300$ samples each, and the ELF-M and ELF-L rows $1{,}000$ at ELF's 64-step preset ($\gamma{=}1.0$). The two final-step sampling rows come from a separate ELF-B batch, and the $1{,}000$-sample default batch of Table~\ref{tab:buckets} reads a tail of $3.85$. The ELF-B 64-step row also changes the SDE churn from $\gamma{=}1.5$ to $1.0$, and dashes mark quantities not scored for a configuration. Generative perplexity improves monotonically along the step axis while repetition rises $12\times$, and no configuration brings the tail near the human rate.}
\label{tab:appsweep}
\begin{tabular}{lrrr}
\toprule
Configuration & tail $z<4$ (\%) & gen.\ perplexity & 4-gram rep.\ (\%) \\
\midrule
ELF-B, 16 steps, $g{=}3$            & 7.05 & 44.51 & 0.63 \\
ELF-B, 32 steps, $g{=}1$            & 3.88 & 29.23 & 1.21 \\
ELF-B, 32 steps, $g{=}3$ (default)  & 3.75 & 24.18 & 2.84 \\
ELF-B, 32 steps, $g{=}5$            & 4.02 & 23.40 & 2.87 \\
ELF-B, 64 steps, $g{=}3$            & 4.57 & 18.97 & 7.61 \\
ELF-B, final-step sampling $T{=}1.0$ & 4.72 & --- & --- \\
ELF-B, final-step sampling $T{=}2.0$ & 4.83 & --- & --- \\
ELF-M, 64 steps                     & 5.85 & 21.60 & --- \\
ELF-L, 64 steps                     & 6.87 & 23.77 & --- \\
\midrule
Human (OpenWebText reference)               & \textbf{11.49} & --- & --- \\
\bottomrule
\end{tabular}
\end{table}

\textbf{What the knobs reach.} Across the full sweep, the tail spans $3.75$--$7.05\%$. ELF-L at the 32-step setting of Table~\ref{tab:main}, a probe-pipeline run (Appendix~\ref{app:configs}), reads $7.42\%$ on seed $31337$ ($200$ samples), and the ELF range of Figure~\ref{fig:fig1}b spans the sweep and this reading, $0.33$--$0.65\times$ the human rate. The range takes ELF-L from seed $31337$ alone: the other two ELF-L stock seeds of Table~\ref{tab:main} read $8.16\%$ and $6.89\%$ under the same rule, so across seeds the upper end reaches $0.71\times$, whereas the LangFlow range spans all four of its stock seeds (below). The sweep's own upper end, the $16$-step configuration, owes most of its tail to samples that fail the non-word bar. Under the dual bar, it accepts $73.0\%$ of its samples, and those carry a tail of $3.99\%$, against $83.7\%$ and $3.56\%$ at the default $32$ steps. Its rejected samples, all rejected for non-words, hold $27\%$ of its words but $59\%$ of its rare ones. No configuration reaches human.

\textbf{No decode-time knob helps.} Nor can the final step recover what the trajectory has lost. By then, the distribution has an effective $1.02$--$1.05$ candidates, measured as the mean over positions of the exponential of its entropy, and $95$--$98\%$ of positions put more than $0.99$ on their top word, so the final choice is nearly determined. Switching from argmax to sampling and then raising the temperature to $2.0$ moves the tail by only $0.13$ percentage points ($4.70\to4.72\to4.83\%$). Whatever is to be decided has been decided before the decoding head is read, which is the claim that \S\ref{sec:cause} then localizes.

\textbf{The same on LangFlow.} LangFlow exposes a step count rather than a guidance scale, so its sweep runs along that axis: $32$, $64$, $128$, and $256$ steps, $n{=}200$ each, read on the accepted set under the dual bar. The human rate there is $11.17\%$, so the ratios below are on that footing rather than on the unfiltered one used above. The accepted-set tail is flat across the axis ($8.54$, $8.17$, $8.42$, and $8.24$, and $7.75$ on a second seed at $256$ steps), while the accept rate climbs monotonically ($34.0$, $50.0$, $55.0$, and $58.5\%$). Steps thus buy coherence but not tail, and the ceiling sits at $0.69$--$0.76\times$ human. Read instead on unfiltered output under the alphabetic rule, the convention that Figure~\ref{fig:fig1}b uses throughout, the same axis gives $9.43$, $9.06$, $9.03$, and $8.71\%$ against the human $11.49\%$ ($8.43\%$ on a second seed at 256 steps). The four stock seeds at the default step count give $8.25$--$8.79\%$, and the LangFlow range of that figure is the span of these readings, $0.72$--$0.82\times$. The fabricated-name signature is stable across the axis, with $60$--$69\%$ of non-words capitalized, the same family of artifact as in \S\ref{sec:collapse}.

\textbf{Where the \graft{} ranges of Figure~\ref{fig:fig1}b come from.} Under the same rule, on unfiltered output, the \graft{} arms on ELF-B, ELF-M, and ELF-L read $9.82$, $12.44$, and $13.21\%$ at the fixed dose of Table~\ref{tab:main}, and $11.94$, $13.18$, and $15.64\%$ at the tuned dose of Appendix~\ref{app:heldout}, both on seed $31337$. Their other two fixed-dose seeds read $9.93$ and $9.49\%$, $12.47$ and $12.10\%$, and $13.03$ and $12.75\%$, respectively. On LangFlow, the arms read $15.45$, $13.94$, and $13.65\%$ at the fixed dose (seeds $42$, $1042$, and $2042$) and $11.77$, $11.49$, and $11.32\%$ at the tuned dose (same seeds). The right end of each \graft{} range is the three-seed mean of the fixed dose on the model that reaches furthest: $13.00\%$ on ELF-L, or $1.13\times$ human, and $14.35\%$ on LangFlow, or $1.25\times$. Single seeds span $1.11$--$1.15\times$ on ELF-L and $1.19$--$1.34\times$ on LangFlow. The left end of each range is the family's own lower bound, which \graft{} inherits because a dose of zero leaves the stock sampler untouched. These readings use the alphabetic rule without the foreign discount, whereas Table~\ref{tab:main} uses the all-token rule with the discount, and neither applies an acceptance filter. The figure accepts this mismatch in exchange for putting all six ranges on one axis.

\textbf{What generative perplexity reads.} Along the step axis ($16\to32\to64$ steps at guidance 3), generative perplexity improves monotonically, from $44.51$ to $24.18$ and $18.97$, while 4-gram repetition rises from $0.63$ to $2.84$ and $7.61\%$, a factor of $12.1$ across the same range. Lowering the guidance to 1 at 32 steps sits in between ($29.23$, $1.21\%$). The metric thus reports a $2.3\times$ improvement over a range on which the text becomes $12\times$ more repetitive. The 64-step point also lowers the SDE churn, but the 16-to-32-step half of the axis keeps the churn fixed and shows the same pattern: generative perplexity improves from $44.51$ to $24.18$, while repetition rises $4.5\times$. This reproduces on our own pipeline what \citet{lpir2026} report for low-diversity text, and it is why \S\ref{sec:collapse} declines to read generative perplexity as a quality proxy.

\subsection{The $q$-decile test and the loss-prediction signature}
\label{app:qdecile}

This subsection gives the measurement behind Finding~3 of \S\ref{sec:cause}: if the denoising head averages a rare word with a stronger competitor, the competitor's probability should predict whether the word is lost. We noise the first $200$ documents of the human reference to $t{=}0.31$, part-way through a generation. We take the positions whose true word is rare, together with a thinned set of common positions as a reference, and keep positions at least twelve tokens apart. On ELF-B, this leaves $12{,}370$ positions: $4{,}036$ of the $15{,}035$ rare positions and $8{,}334$ common ones. At each position, we record whether the original word survives and $q$, the probability the decoding head assigns to the strongest candidate other than the original word.

Sorting the rare positions by $q$ into ten equal groups gives loss rates of $7$, $20$, $45$, $78$, $84$, $88$, $86$, $85$, $90$, and $92\%$, the curve reported in \S\ref{sec:cause}. The curve rises steeply over the first four groups, from $7$ to $78\%$, and then levels off at $84$--$92\%$ without rising strictly from group to group, which is the shape a threshold produces rather than a gradual bias. Over all positions, the same curve reads $1$, $3$, $7$, $19$, $57$, $73$, $78$, $79$, $81$, and $88\%$. As a predictor of loss, $q$ reaches an AUC of $0.842$ on the rare positions and $0.880$ over all positions. Of the rare positions, $67.5\%$ lose their word, against $39.3\%$ of the common positions. Moreover, when a rare word is lost, the strongest competitor takes its place $31.8\%$ of the time ($n{=}2{,}724$), against a chance level of $0.003\%$ for any single word in the $32$k vocabulary.

The signature holds on every model we measured, including the small ELF-S models of Appendix~\ref{app:training}, where it is weaker. Over all positions, the AUC and the rate at which the competitor replaces a lost word are $0.880$ and $39.3\%$ on ELF-B, $0.832$ and $25.6\%$ on the ELF-S base arm, and $0.825$ and $28.1\%$ on the ELF-S infill arm; on rare positions alone, the AUC falls from $0.842$ to $0.731$ and $0.689$. Weaker decoding heads thus give a weaker signature and gain less from committing rare words (Table~\ref{tab:appscale}). Per-position records for the ELF-B measurements will be released with the code.

\subsection{Foreign-dominant vocabulary and the discounted tail}
\label{app:foreign}

The tail is defined by English Zipf frequency, so a word that is rare in English counts toward it even when it is ordinary in another language. This subsection details the language gate that keeps such words out of \graft{}'s commitments and the foreign discount that removes them from the measured tail, and it reports how much of each arm's gain they carry.

\textbf{The gate.} A word is foreign-dominant at a given margin when its highest Zipf frequency across thirteen other languages exceeds its English Zipf by more than that margin. The languages are German, French, Romanian, Spanish, Italian, Dutch, Portuguese, Polish, Catalan, Turkish, Indonesian, Swedish, and Danish, all read from the same \texttt{wordfreq} lexicon as the English frequencies (Appendix~\ref{app:metrics}). Words that meet this test at a margin of $2.5$ are struck from the vocabulary that \graft{} may commit. The margin, rather than a binary ``commoner in any language'' rule, is what keeps English-adopted proper nouns available, as the comparison below shows.

\textbf{Why ELF needs it.} ELF's T5 vocabulary supplies such words precisely where \graft{} reaches: of the alphabetic vocabulary entries that are rare in English ($0{<}z{<}4$), $24.3\%$ are foreign-dominant at margin $2.5$ ($25.8\%$ of the word-initial ones), against $3.4\%$ ($2.5\%$) of GPT-2's. Table~\ref{tab:appforeign} therefore decomposes every arm's accepted-set tail into the part carried by foreign-dominant words and the remainder, using the tokenization, dual bar, pooling, and bootstrap of the accept-then-measure protocol (Table~\ref{tab:appaccept}) and discounting the human reference in the same way.

\textbf{What the decomposition shows.} Two regularities hold across the grid. First, the foreign share grows with how deep the Zipf band reaches. Without the language gate, the $(8,5,3)$ quota at $q_{\min}{=}0.02$ carries $1.8$ to $4\times$ the share of the fixed dose at $q_{\min}{=}0.08$ on every ELF scale. With the gate, the ratio is smaller, $1.75\times$ on ELF-M and $1.25\times$ on ELF-L; on ELF-B it reverses ($2.1\%$ against $5.0\%$), but that pair also differs in step schedule. Second, the share is worst where the decoding head is weakest, which puts ELF-B's fixed dose and ELF-L's tuned arm at comparable contamination from opposite directions. At a margin of $2.5$, the foreign function words score $+2.9$ to $+3.7$ above their English Zipf frequency, against $-0.6$ to $+1.9$ for English-adopted proper nouns such as \emph{Munich}, \emph{Bayern}, and \emph{Tottenham}, which a binary rule (any language higher) would delete. The language gate thus removes most of the foreign share at almost no cost in raw gain. On ELF-L's tuned arm, the two readings cross: the gate lowers the raw gain from $+9.32$ to $+8.41$ while raising the discounted gain from $+7.59$ to $+8.32$. LangFlow's fixed dose needs no such correction, since its foreign share ($1.1\%$) matches stock and human. Its band quotas, which reach deeper bands, carry more ($2.1$--$5.9\%$ of the accepted-set tail, Table~\ref{tab:applfdose}), and Table~\ref{tab:appaccept} discounts them. At the same fixed dose, ELF-B's $13.8\%$ against LangFlow's $1.1\%$ is the clearest evidence that the hazard is a property of the tokenizer rather than of the repair.

\begin{table}[ht]
\centering
\small
\caption{Foreign-dominant share of the accepted-set tail, and the gain with and without the foreign discount. All $\Delta$ are against the same-model stock sampler, and every discounted $\Delta$ shown is significant at $p{<}0.001$. \emph{Gated} arms carry both gates of \S\ref{sec:method}; \emph{tuned} is the $(8,5,3)$ band quota at $q_{\min}{=}0.02$, and \emph{fixed} is the fixed dose at $q_{\min}{=}0.08$. ELF rows are seed $31337$, except the ELF-B row \emph{fixed, gated}, which averages three batch-16 runs with both gates; the LangFlow fixed row averages three seeds, and the LangFlow stock row is the seed-$42$ run. The other ELF-B rows differ in their settings: the fixed row is a batch-8 run with the attestation gate and no language gate, and the tuned row is a batch-8 run with the attestation gate and the step-schedule shift. The tuned, gated row is the seed-$31337$ run of the tuned arm of Appendix~\ref{app:heldout} (batch $8$, both gates, default step schedule), so it differs from the tuned row in step schedule as well as in the language gate.}
\label{tab:appforeign}
\begin{tabular}{llrrrr}
\toprule
Model & Arm & \tailacc{} & foreign\,\% & $\Delta$ raw & $\Delta$ discounted \\
\midrule
human    & ---              & 11.17 & 0.8 & --- & --- \\
\addlinespace[2pt]
ELF-B    & stock            &  4.30 & 1.6 & --- & --- \\
         & fixed            &  9.67 & 13.8 & $+5.37$ & $+4.10$ \\
         & fixed, gated     &  9.54 & \textbf{5.0} & $+5.24$ & $\mathbf{+4.83}$ \\
         & tuned            & 12.53 & 24.5 & $+8.23$ & $+5.23$ \\
         & tuned, gated     & 11.65 & \textbf{2.1} & $+7.35$ & $\mathbf{+7.18}$ \\
\addlinespace[2pt]
ELF-M    & stock            &  5.23 & 0.8 & --- & --- \\
         & fixed            & 11.83 & 0.6 & $+6.59$ & $+6.56$ \\
         & fixed, gated     & 11.82 & \textbf{0.4} & $+6.59$ & $\mathbf{+6.58}$ \\
         & tuned            & 12.57 & 2.4 & $+7.33$ & $+7.08$ \\
         & tuned, gated     & 12.28 & \textbf{0.7} & $+7.04$ & $\mathbf{+7.00}$ \\
\addlinespace[2pt]
ELF-L    & stock            &  6.66 & 0.7 & --- & --- \\
         & fixed            & 13.28 & 3.2 & $+6.62$ & $+6.23$ \\
         & fixed, gated     & 12.87 & \textbf{0.8} & $+6.21$ & $\mathbf{+6.15}$ \\
         & tuned            & 15.99 & 11.1 & $+9.32$ & $+7.59$ \\
         & tuned, gated     & 15.08 & \textbf{1.0} & $+8.41$ & $\mathbf{+8.32}$ \\
\addlinespace[2pt]
LangFlow & stock            &  8.28 & 1.3 & --- & --- \\
         & fixed            & 11.83 & 1.1 & $+3.55$ & $+3.54$ \\
\bottomrule
\end{tabular}
\end{table}

\textbf{The argmax controls.} The argmax controls settle on the same measure. On ELF-B, the gated control reads $+0.62$ raw, of which foreign words carry $+0.57$, leaving a discounted $+0.05$ [$-0.39$, $+0.48$]; the ungated control reads $+0.64$ raw and a discounted $+0.12$ [$-0.29$, $+0.54$]. Neither is distinguishable from zero. LangFlow's control carries only the attestation gate, since LangFlow has no language gate, and reads $-0.10$ [$-0.89$, $+0.70$] after the discount. The language gate does not reduce the controls' foreign share, because the gates filter only rare candidates and a control commits the argmax without passing through them. The argmax is usually a common word, but not always: of the gated ELF-B control's $6{,}200$ commitments, $166$ are rare words, $128$ of them foreign-dominant (\emph{das}, \emph{vor}, \emph{sein}, \emph{aber}), and splicing keeps them all. The other foreign words in a control's output were written by the flow.

\textbf{Foreign words in \graft{}'s own output.} Foreign words written by the flow are also what survives in \graft{}'s own output. Counting every alphabetic word that is foreign-dominant at margin $2.5$, including words absent from the English lexicon, they make up $0.79\%$ of alphabetic words ($0.65$--$0.93\%$ across three seeds), against $0.30\%$ for stock.

\subsection{The accept-then-measure protocol, and metrics we do not headline}
\label{app:accept}

Table~\ref{tab:main} reports unfiltered output, so that no filtering step can flatter it. This subsection gives the complementary reading, the accept-then-measure protocol of \S\ref{sec:premises}, reads the argmax controls under both protocols, and reports the two metrics that move against \graft{}.

\textbf{The protocol.} The protocol, its bars, and \tailacc{} are defined in Appendix~\ref{app:metrics}. Table~\ref{tab:appaccept} gives every arm of Table~\ref{tab:main}, together with the tuned rows and the ladder, under this protocol, and, as in Table~\ref{tab:main}, the fixed-dose \graft{} rows stay above stock and, where it is run, best-of-4. Whether the tail was bought at the bars is a question about the accept column. Accept \emph{rises} on ELF-B ($66.0\to68.3\%$) and is flat on ELF-L ($63.5\to63.4\%$), whereas it falls on ELF-M ($81.0\to65.7\%$) and on LangFlow ($60.5\to48.5\%$).

\textbf{The argmax controls.} \S\ref{sec:controls} reads the two argmax controls on the unfiltered output of Table~\ref{tab:main}. For that reading, we resample the samples of each arm ($2{,}000$ resamples, no acceptance filter) and recompute the tail of Table~\ref{tab:main} in every replicate. On ELF-B, the control's tail gain against the stock run of the same seed is $+0.30$ [$-0.13$, $+0.73$] ($p{=}0.17$). On LangFlow, it is $+0.59$ [$-0.25$, $+1.44$] ($p{=}0.18$) against the three-seed stock mean of Table~\ref{tab:main}, and $+0.72$ [$-0.12$, $+1.65$] ($p{=}0.10$) against the stock run of the control's own seed. All three intervals include zero, whereas the same resampling puts the \graft{} gain on ELF-B at $+4.75$ [$+4.41$, $+5.10$]. Under the accept-then-measure protocol, after the foreign discount, the controls read $+0.05$ [$-0.39$, $+0.48$] on ELF-B and $-0.10$ [$-0.89$, $+0.70$] on LangFlow (Appendix~\ref{app:foreign}). The per-seed values behind the three-seed rows of Table~\ref{tab:main} are given in Table~\ref{tab:appperseed}.

\begin{table}[ht]
\centering
\small
\caption{Accept-then-measure protocol ($n{=}200$ per arm; the human reference reads a \tailacc{} of $11.17$ at an accept rate of $90.5\%$). In every row, $\Delta$ is against the same-model stock sampler after the foreign discount (Appendix~\ref{app:foreign}). The \graft{} rows, the tuned rows, and the ELF-B ladder row average three seeds ($31337$, $55555$, and $77777$ on ELF; $42$, $1042$, and $2042$ on LangFlow), with each seed read against the same stock run. The stock, argmax-control, and LangFlow ladder rows are single runs, and each best-of-4 row selects among four stock runs. Every $\Delta$ is significant at $p{<}0.001$ on every seed, except those of the two argmax controls and of LangFlow's tuned row ($p{=}0.29$, $0.017$, and $0.029$ by seed). The ladder rows predate the gates and are ungated, and the ELF-B tuned row runs without the step-schedule shift (Appendix~\ref{app:heldout}). Among the \graft{} and ladder rows, bold marks an accept rate at or above the same-model stock rate. The ladder is a per-sample controller over ungated \graft{} variants, not \graft{} itself (Appendix~\ref{app:configs}).}
\label{tab:appaccept}
\begin{tabular}{lrrr}
\toprule
Sampler & accept\,\% & \tailacc{}\,\% & $\Delta$\tailacc{} \\
\midrule
\multicolumn{4}{l}{\emph{ELF-B}}\\
stock                & 66.0 & 4.30 & --- \\
stock best-of-4      & 98.0 & 5.64 & $+1.3$ \\
argmax control       & 74.5 & 4.92 & $+0.05$ (n.s.) \\
\graft{}             & \textbf{68.3} & 9.54 & $+4.83$ \\
\graft{} (tuned)     & 57.3 & 11.74 & $+7.22$ \\
ladder               & \textbf{88.5} & 9.23 & $+3.8$ \\
\addlinespace[2pt]
\multicolumn{4}{l}{\emph{ELF-M}}\\
stock                & 81.0 & 5.23 & --- \\
\graft{}             & 65.7 & 11.61 & $+6.35$ \\
\addlinespace[2pt]
\multicolumn{4}{l}{\emph{ELF-L}}\\
stock                & 63.5 & 6.66 & --- \\
\graft{}             & 63.4 & 12.47 & $+5.72$ \\
\graft{} (tuned)     & \textbf{65.6} & 14.71 & $+7.91$ \\
\addlinespace[2pt]
\multicolumn{4}{l}{\emph{LangFlow}}\\
stock                & 60.5 & 8.28 & --- \\
stock best-of-4      & 99.0 & 9.73 & $+1.4$ \\
argmax control       & 47.0 & 8.12 & $-0.10$ (n.s.) \\
\graft{}             & 48.5 & 11.83 & $+3.54$ \\
\graft{} (tuned)     & 50.3 & 9.24 & $+0.76$ \\
ladder               & \textbf{84.5} & 12.79 & $+4.4$ \\
\bottomrule
\end{tabular}
\end{table}

\textbf{Metrics we do not headline.} Two metrics move against \graft{}, and both rely on GPT-2-large: clean-PPL on its likelihood, and MAUVE \citep{mauve2021} on its features. Clean-PPL rises from $26.5$ to $46.8$--$66.1$ across the ELF-B dose axis (Appendix~\ref{app:buckets}). MAUVE, computed with the standard GPT-2-large featurizer at $n{=}200$ on ELF-B, ELF-L, and LangFlow (ELF-M was not measured), falls on each of these models, and it falls monotonically with dose: from $0.851$ to $0.777$ at the fixed dose and to $0.640$ at the tuned dose on ELF-B, from $0.915$ to $0.799$ and $0.756$ on ELF-L, and from $0.927$ to $0.803$ on LangFlow. We do not headline these metrics because the claim of this paper is lexical, namely that the tail is missing and can be restored, not that the text is better. Clean-PPL takes its notion of ``human-like'' from GPT-2 likelihood, which scores rare words as unlikely by construction; this is the same bias that makes generative perplexity reward the defect we set out to measure (\S\ref{sec:collapse}), and the human reference itself carries an $11.2\%$ tail at a clean-PPL of $16.66$. MAUVE compares distributions of GPT-2-large features instead, so this argument does not apply to it directly. We report these numbers here rather than omit them, and we regard blind human preference, which we did not run at scale, as the test that would settle the quality question in either direction.

\subsection{Held-out dose selection}
\label{app:heldout}

Table~\ref{tab:main} uses one fixed dose on all three ELF models and a LangFlow-specific dose on LangFlow, where the fixed dose overshoots (\S\ref{sec:dose}). This subsection reports how that dose was selected, and what a tuned dose would buy on every model.

\textbf{Selection on LangFlow.} A tuned dose invites the objection that it was chosen on the samples it is read on. On LangFlow, the one model where Table~\ref{tab:main} uses such a dose, we therefore select and report on disjoint seeds. Six candidate doses were ranked by JSD (Appendix~\ref{app:metrics}) on seed $42$. The winner, the $(0,1,1)$ quota with the attestation gate, is read on seeds $1042$ and $2042$, and its runs on those seeds played no part in the ranking. Seed $1042$ did enter the ranking once, through a losing candidate (the $z{<}3$ band at $q_{\min}{=}0.02$, ``\graft{}, deeper'' in Table~\ref{tab:applfdose}), which was ranked on its seed-$1042$ run.

\textbf{No per-model selection on ELF.} On ELF, the dose was not selected per model. The $(8,5,3)$ quota was compared with three re-balanced quotas on ELF-B only (seed $31337$), by the JSD of the accepted set and in an earlier configuration (batch $8$, an earlier version of the language gate that excluded only German-dominant words, no attestation gate, and no foreign discount). ELF-M and ELF-L reuse it unchanged. On ELF-B, the two doses compared below also differ in sampler settings: the tuned arms run with the default step schedule and batch $8$, whereas the fixed-dose arms use the step-schedule shift and batch $16$, so the ELF-B comparison mixes the dose with the step schedule. On ELF-M and ELF-L, neither dose uses the step-schedule shift.

\textbf{What a tuned dose buys.} Table~\ref{tab:appheldout} reads the fixed and the tuned dose on the same three seeds for every model. The tuned dose wins on all three seeds of all four models, twelve comparisons out of twelve, and what differs between the models is the size of the win. It is about fourfold on ELF-B ($4.0$--$4.5\times$ per seed) and $16$--$142\times$ per seed on LangFlow, where the largest ratio comes from a tuned JSD of $0.00005$ and the three-seed means differ by $34\times$. On ELF-M and ELF-L, the win is only $1.04$--$1.40\times$. Their fixed dose already lands on the human tail, and the tuned dose improves the JSD while overshooting the tail ($1.09\times$ and $1.27\times$, against the fixed dose's $1.01\times$ and $1.10\times$). This is why Table~\ref{tab:main} keeps the fixed dose on ELF and uses the tuned dose only on LangFlow.

\begin{table}[ht]
\centering
\small
\caption{Held-out dose selection: JSD to the human profile on unfiltered output. \emph{Fixed} is the fixed dose, $m{=}8$ at $q_{\min}{=}0.08$, and \emph{tuned} is the band quota, $(8,5,3)$ on ELF and $(0,1,1)$ on LangFlow, both at $q_{\min}{=}0.02$. On LangFlow, the winner was chosen on the \emph{selection} seed ($42$), and its \emph{held-out} runs ($1042$ and $2042$) played no part in the choice, although one losing candidate was ranked on seed $1042$ (see text). On ELF, the columns are seeds $31337$, $55555$, and $77777$; the quota was chosen on ELF-B under an earlier configuration and reused on ELF-M and ELF-L, and on ELF-B the tuned arms also run without the step-schedule shift (see text). The tuned dose gives the lower JSD on every seed of every model.}
\label{tab:appheldout}
\begin{tabular}{@{}llrrr@{}}
\toprule
Model & Dose & selection & held-out 1 & held-out 2 \\
\midrule
ELF-B    & fixed & .0049 & .0047 & .0052 \\
         & tuned   & \textbf{.0012} & \textbf{.0012} & \textbf{.0012} \\
\addlinespace[2pt]
ELF-M    & fixed & .0082 & .0083 & .0072 \\
         & tuned   & \textbf{.0059} & \textbf{.0079} & \textbf{.0052} \\
\addlinespace[2pt]
ELF-L    & fixed & .0049 & .0048 & .0041 \\
         & tuned   & \textbf{.0047} & \textbf{.0045} & \textbf{.0033} \\
\addlinespace[2pt]
LangFlow & fixed & .0075 & .0044 & .0055 \\
         & tuned   & \textbf{.0001} & \textbf{.0003} & \textbf{.0002} \\
\bottomrule
\end{tabular}
\end{table}

\textbf{A frozen configuration on new seeds.} Seeds used in development could flatter a configuration chosen on them. We therefore froze the champion$+$co-pinning configuration, with the anti-repetition correction and, on ELF-B, the step-schedule shift, and ran it unchanged on seeds that no earlier run had used. Each seed was read against a stock run on the same seed, and no run was selected or dropped. Readings use the dual bar and the accepted-set tail, without the foreign discount. On ELF-B, seeds $20260911$, $20260912$, and $20260913$ give gains of $+5.22$, $+5.55$, and $+5.19$, and $+5.28$, $+5.52$, and $+5.12$ with the attestation gate added; no lower CI bound is below $+4.57$, and the three seeds on which the configuration was developed gave $+5.20$ to $+5.39$. On LangFlow, seed $90011$ with the attestation gate gives $+4.43$ [$+2.12$, $+7.01$], and seed $90012$ without it gives $+3.95$ [$+2.05$, $+6.15$]. Both lie inside the development range of $+3.07$ to $+5.33$ (Table~\ref{tab:appmcb}).

\subsection{The LangFlow dose sweep: profile and tail cannot be had together}
\label{app:lfdose}

This subsection explains why the distance-to-human-profile columns of Table~\ref{tab:main} read as they do on LangFlow. Table~\ref{tab:applfdose} orders the LangFlow arms by dose. As the dose rises from the lowest setting we ran, the JSD and the tail both rise monotonically, so the Zipf profile moves away from human as the tail grows.

\textbf{The lowest dose.} The lowest dose, which Table~\ref{tab:main} reports as \graft{} (tuned), matches the human Zipf profile to a JSD of $.0001$ on seed $42$, the closest reading anywhere in this paper. Its accepted-set tail gain is small: $+0.60$pp on seed $42$, where it is not significant ($p{=}0.16$), and $+1.23$ [$+0.14$, $+2.52$] ($p{=}0.015$) and $+1.08$ [$+0.23$, $+1.95$] ($p{=}0.012$) on the two held-out seeds. Because this quota reaches deeper bands, where foreign-dominant words are commoner, those words carry $3.3$--$3.6\%$ of its accepted-set tail, against stock's $1.3\%$; after the foreign discount, the three readings are $+0.41$, $+0.99$, and $+0.88$ (Table~\ref{tab:appaccept}). On unfiltered output, the dose carries the tail from $0.77$ to $1.03\times$ human (three seeds, Table~\ref{tab:main}).

\textbf{The fixed dose and best-of-4.} The fixed dose buys $+4.80$pp of accepted-set tail on seed $42$, at a JSD $3\times$ that of stock. Over its three seeds, the gain is $+3.55$pp ($+3.54$ after the foreign discount, Table~\ref{tab:appaccept}), and the JSD is $.0058$, $2.3\times$ stock's. The equal-budget stock best-of-4, by contrast, stays close to the human profile on both statistics (JSD $.0015$, $|\Delta\zeta|$ $.002$) while buying $+1.45$pp of accepted-set tail. It is closer than every \graft{} arm on $|\Delta\zeta|$, and closer than all but the two lowest doses on JSD. None of this holds on ELF, where stock sits far from the human profile and leaves room to move toward it. Table~\ref{tab:applfquota} gives the band-quota sweep behind the same conclusion, and Table~\ref{tab:appmcb} gives the equal-NFE comparison with MCB.

\begin{figure}[ht]
\centering
\includegraphics[width=0.62\linewidth]{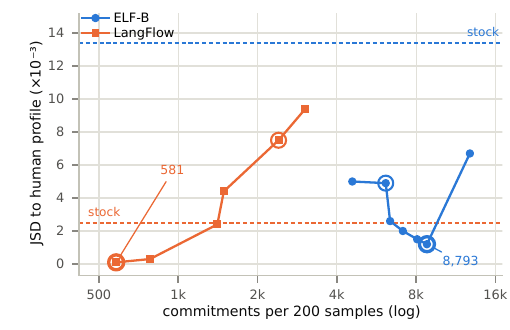}
\caption{JSD to the human profile against dose: ELF-B is U-shaped, LangFlow rises throughout, and the two optima sit $15\times$ apart. Doses are commitments per $200$ samples, read on unfiltered output from single-seed sweeps. On ELF-B, the minimum lies at $8{,}793$ commitments, and every dose we ran improves on the stock sampler. On LangFlow, the JSD rises over the whole range we ran, so the lowest dose, $581$, is the best, and the fixed dose ($2{,}403$) lands \emph{above} the stock sampler. Open rings mark the fixed dose of Table~\ref{tab:main} and the tuned dose of Appendix~\ref{app:heldout}, whose three-seed means are $.0049$ and $.0012$ on ELF-B and $.0058$ and $.0002$ on LangFlow, and dashed lines mark the stock samplers. The ELF-B points mix settings (see text).}
\label{fig:dose}
\end{figure}

\textbf{The two models side by side.} Figure~\ref{fig:dose} plots the same JSD against dose on both models. On LangFlow, the fixed dose lands above its own stock sampler, so tuning there repairs an overshoot that the fixed dose itself creates. The ELF-B sweep in the figure (seed $31337$) combines arms run under different settings. The four intermediate quotas use batch $8$ with the earlier German-only language gate and no attestation gate. The fixed-dose ($6{,}132$) and tuned ($8{,}793$) points carry both gates, and the 16-per-round point ($12{,}800$) carries the attestation gate only. Every ELF-B point uses the step-schedule shift except the tuned one, which runs with the default step schedule, so the minimum of the U-shaped ELF-B curve also differs from its neighbors in step schedule.

\begin{table}[ht]
\centering
\small
\caption{LangFlow dose sweep on unfiltered output, ordered by commitments per $200$ samples. Every row is a single run on seed $42$, except the best-of-4 row, which selects among four stock runs, and \graft{}, deeper, which is seed $1042$. \graft{}, deeper is the $z{<}3$ band at $q_{\min}{=}0.02$; it and the rows marked \emph{ungated} run without the attestation gate. The last two columns give the accepted-set gain against the LangFlow stock sampler, raw and after the foreign discount (Appendix~\ref{app:foreign}). The foreign share of the accepted-set tail is $1.2$--$1.4\%$ for stock, best-of-4, and \graft{}, but $2.1$--$5.9\%$ for the quotas and \graft{}, deeper, which reach deeper bands. For comparison, Table~\ref{tab:main} reports three-seed means for stock ($|\Delta\zeta|$ $.011$), \graft{} ($.0058$, $.063$), and \graft{} (tuned) ($.0002$, $.045$).}
\label{tab:applfdose}
\begin{tabular}{lrrrrrr}
\toprule
Arm & commits & JSD$\,\downarrow$ & $|\Delta\zeta|\,\downarrow$ & accept\,\% & $\Delta$\tailacc{} & disc. \\
\midrule
stock                & --- & .0025 & .023 & 60.5 & --- & --- \\
stock best-of-4      & --- & .0015 & \textbf{.002} & 99.0 & $+1.45$ & $+1.42$ \\
quota $(0,1,1)$      & 581 & \textbf{.0001} & .040 & 51.3 & $+0.60$ (n.s.) & $+0.41$ \\
quota $(0,1,1)$ ungated & 781 & .0003 & .052 & 49.5 & $+0.68$ (n.s.) & $+0.61$ \\
quota $(1,1,0)$ ungated & 1407 & .0024 & .079 & 42.4 & $+2.49$ & $+2.30$ \\
quota $(1,1,1)$      & 1489 & .0044 & .091 & 39.2 & $+2.89$ & $+2.46$ \\
\graft{}             & 2403 & .0075 & .058 & 52.5 & $\mathbf{+4.80}$ & $+4.76$ \\
\graft{}, deeper     & 3035 & .0094 & .054 & 44.5 & $+6.98$ & $+6.19$ \\
\bottomrule
\end{tabular}
\end{table}

\subsection{Zipf profiles, JSD, and the attestation audit}
\label{app:buckets}

Three questions decide whether a method that raises a frequency statistic is worth anything: how much of the gain is dose, where the gain lands, and what it costs. Table~\ref{tab:appdose} addresses them on ELF-B.

\textbf{How much of the gain is dose.} Doubling the dose to $m{=}16$ takes $\Delta$\tailacc{} from $+5.4$ to $+10.1$pp \emph{while accept rises} to $74.0\%$ ($76.5\%$ with a stricter attestation lexicon built from $500$ documents). Champion$+$co-pinning is therefore a conservative operating point rather than a ceiling. Splitting the same nominal dose ($16$ per round) across Zipf bands does not buy more tail: the dose $m{=}16$ beats the $(8,5,3)$ quota, which also fills fewer positions ($9{,}517$ commitments against $12{,}799$).

\textbf{Where the gain lands.} The split instead moves the gain into the band that champion$+$co-pinning cannot reach. The $[2,3)$ band rises from $1.06$ to $2.40$, statistically indistinguishable from the human $2.41$, whereas the dose $m{=}16$ overshoots $[3,4)$. JSD to the human profile, computed on unfiltered output so that acceptance cannot flatter it, orders the arms in the same way. On the runs of Table~\ref{tab:appbuckets}, it is $0.0124$ for stock, $0.0032$ for champion$+$co-pinning, $0.0067$ for dose $16$, and $0.0011$ for the quota; the dose-16 and quota runs there carry the earlier German-only language gate and no attestation gate, and the attestation-gated dose-16 run of Table~\ref{tab:appdose} reads $0.0058$. A third knob, narrowing the candidate band to Zipf ${<}3$, trades total gain for depth: it reaches $2.71\%$ of the unfiltered output in $[2,3)$ and $1.21\%$ below $z{=}2$, at $+3.15$pp of accepted-set tail. No configuration we tested matches human across all bands at once.

\textbf{What the gain costs.} The three samplers nonetheless pay the same price. Clean-PPL rises from $26.5$ for stock to $46.8$ for champion$+$co-pinning ($+5.34$), $56.7$ for the $(8,5,3)$ quota ($+7.73$), and $66.1$ for dose $16$ ($+9.91$, its ungated run in Table~\ref{tab:appattest}). These are $0.26$, $0.26$, and $0.25$pp of tail per point of clean-PPL, three points on one line. The gain is coverage rather than repetition: rare types per $100$k, the types/100k count restricted to rare words, rise from $2{,}124$ to $3{,}421$, against a human $3{,}985$, with $27.5$--$55.3$ distinct committed types per sample.

\textbf{Batch effects, and where the rest is.} A batch control, which reruns the same configurations with the batch sizes $8$ and $16$ exchanged, moves $\Delta$\tailacc{} by only $0.05$--$0.21$pp while accept moves by $1.5$--$4$pp, which confirms that accept rates, not tail deltas, carry the volatility of this protocol. Table~\ref{tab:appbuckets} gives the Zipf profiles and JSDs, and Table~\ref{tab:appattest} gives the attestation audit behind the gate. The other gate, the language gate, is ablated in Appendix~\ref{app:foreign} (Table~\ref{tab:appforeign}). The GPT-2-large appropriateness judge cited in \S\ref{sec:premises} is described in Appendix~\ref{app:checks}, and its re-scoring in the selection-time draft context is reported in Appendix~\ref{app:heads}.

\begin{table}[ht]
\centering
\small
\caption{Dose and placement on ELF-B, all against a batch-matched stock sampler, the stock run at the same batch size and seed (accept $62.0\%$, \tailacc{} $4.31$; see the batch control in the text). The four band columns are accepted-set shares with bootstrap CIs against human, and bold marks bands statistically indistinguishable from human. The band-quota and dose-16 arms without the attestation gate ran with an earlier German-only language gate, which their gated counterparts drop, so only the champion$+$co-pinning pair isolates the attestation gate (Table~\ref{tab:appattest}). The dose-16 and band-quota arms select at $q_{\min}{=}0.02$, against $0.08$ for champion$+$co-pinning. The three samplers differ in where the tail lands, not in what it costs: the tail bought per point of clean-PPL is $0.25$--$0.26$ for all of them.}
\label{tab:appdose}
\begin{tabular}{lrrrrrrr}
\toprule
Sampler & accept\,\% & \tailacc{} & $\Delta$ & $z{\geq}4$ & $[3,4)$ & $[2,3)$ & $z{<}2$ \\
\midrule
Human reference & 90.5 & 11.17 & --- & 88.39 & 8.01 & 2.41 & 0.74 \\
Stock           & 62.0 &  4.31 & --- & 94.35 & 3.53 & 0.50 & 0.29 \\
champion$+$co-pinning ($m{=}8$) & 68.0 &  9.65 & $+5.34$ & \textbf{88.71} & \textbf{8.01} & 1.06 & 0.58 \\
\quad $+$ attestation gate       & 61.5 &  9.67 & $+5.36$ & \textbf{88.68} & \textbf{8.15} & 1.01 & 0.51 \\
Dose $m{=}16$ $+$ gate           & 74.0 & 14.39 & $+10.08$ & 84.06 & 12.40 & 1.34 & \textbf{0.66} \\
Band quota $(8,5,3)$           & 72.0 & 12.04 & $+7.73$ & 86.50 & 8.54 & \textbf{2.40} & 1.10 \\
\quad $+$ attestation gate       & 72.0 & 12.53 & $+8.22$ & 85.95 & 8.46 & 2.93 & 1.13 \\
\bottomrule
\end{tabular}
\end{table}

\begin{table}[ht]
\centering
\small
\caption{Zipf profiles on \emph{unfiltered} output, shown for the four rarer bands, with JSD to the human profile. As in Appendix~\ref{app:metrics}, band shares are over lexicon words, with non-words removed and the shares renormalized; the non-word column is over all word tokens. Tables~\ref{tab:buckets} and~\ref{tab:appunfiltered} keep non-words in the denominator and strip slightly more punctuation, so their band shares differ slightly (human $[4,5)$: $21.27$ there, $21.34$ here). The ELF-B rows are batch-8 runs on seed $31337$, and every JSD in this table is computed without the foreign discount. The stock, champion$+$co-pinning, and $(8,5,3)$ rows are the runs of Table~\ref{tab:appdose} without the attestation gate; the $m{=}16$ row is the dose-16 run without that gate (Table~\ref{tab:appattest}), not the gated run of Table~\ref{tab:appdose}, which reads $.0058$; and the $m{=}8$, $q_{\min}{=}0.02$ and $(6,5,2)$ rows appear only here. Every ELF-B row except stock and champion$+$co-pinning carries the earlier German-only language gate and selects at $q_{\min}{=}0.02$. The stock row reads $.0124$ here against $.0134$ in Table~\ref{tab:main} mainly because the two stock runs differ (batch $8$ here, batch $16$ there): without the discount, the batch-16 run reads $.0135$, so the discount itself moves it by only $.0001$. On ELF-B, champion$+$co-pinning matches the common bands and $[3,4)$ but leaves $[2,3)$ at under half the human rate, a doubled dose overshoots $[3,4)$, and the two band quotas are the only arms that reach $[2,3)$, at the cost of depleting the common bands.}
\label{tab:appbuckets}
\begin{tabular}{lrrrrrr}
\toprule
ELF-B arm & $[4,5)$ & $[3,4)$ & $[2,3)$ & $z{<}2$ & non-word & JSD \\
\midrule
Human                  & 21.34 & 8.13 & 2.52 & 0.86 & 0.66 & --- \\
Stock                  & 22.88 & 3.76 & 0.65 & 0.35 & 2.40 & 0.0124 \\
Champion$+$co-pinning ($m{=}8$) & 20.24 & 8.23 & 1.12 & 0.64 & 2.30 & 0.0032 \\
$m{=}8$, $q_{\min}{=}0.02$ & 20.39 & 8.52 & 0.98 & 0.57 & 2.24 & 0.0038 \\
$m{=}16$               & 18.03 & 12.79 & 1.23 & 0.74 & 2.01 & 0.0067 \\
Band quota $(8,5,3)$ & 19.22 & 8.78 & 2.55 & 1.19 & 2.00 & \textbf{0.0011} \\
Quota $(6,5,2)$        & 19.78 & 7.62 & 2.55 & 1.13 & 2.05 & 0.0011 \\
\midrule
LangFlow arm & & & & & & \\
Human                  & 21.34 & 8.13 & 2.52 & 0.86 & 0.66 & --- \\
Stock                  & 20.98 & 7.08 & 1.27 & 0.43 & 2.44 & 0.0026 \\
Champion$+$co-pinning   & 21.12 & 14.74 & 1.47 & 0.43 & 2.41 & 0.0098 \\
Quota $(1,2,3)$        & 22.53 & 11.80 & 5.07 & 2.39 & --- & 0.0117 \\
\bottomrule
\end{tabular}
\end{table}

\begin{table}[ht]
\centering
\small
\caption{Attestation audit. A commitment is \emph{unattested} when the lowercased form of its token never appears in 30k OpenWebText documents (documents $1{,}000$--$30{,}999$ of the reference shard, disjoint from the human reference and hence from the bar calibration). Contraction fragments (\emph{wouldn}, \emph{couldn}), an artifact of whitespace tokenization in the lexicon builder, make up $10$--$13\%$ of the unattested counts and are excluded from the numerator of the percentages, whose denominator is all commitments. The gate drives the unattested count to zero in every arm. On ELF-B, only the champion$+$co-pinning pair isolates the gate, since the other ELF-B pairs also differ in a language gate (see text).}
\label{tab:appattest}
\begin{tabular}{llrrrr}
\toprule
Model & Arm & commits & unattested & $\Delta$\tailacc{} & with gate \\
\midrule
ELF-B & champion$+$co-pinning & 6{,}220 & $2.1\%$ & $+5.34$ & $+5.36$ \\
ELF-B & band quota          & 9{,}517 & $7.8\%$ & $+7.73$ & $+8.22$ \\
ELF-B & dose $m{=}16$       & 12{,}800 & --- & $+9.91$ & $+10.08$ \\
LangFlow & champion$+$co-pinning & 2{,}429 & $0.8\%$ & $+5.33$ & $+4.80$ \\
LangFlow & quota $(1,2,3)$  & 2{,}640 & $22.1\%$ & $+8.41$ & $+6.77$ \\
\bottomrule
\end{tabular}
\end{table}

\begin{table}[ht]
\centering
\small
\caption{LangFlow dose response of the band quota, with quotas given as commitments per round in each band. Every setting we tried overshoots. The realized commitment count is limited by supply rather than by the quota: the fill rate, the share of the quota's commitment slots that are actually filled, is $54$--$55\%$ across these arms and is set by $q_{\min}$, not by the quota, so narrowing the band while lowering $q_{\min}$ \emph{raises} commitments from $2{,}429$ to $3{,}345$.}
\label{tab:applfquota}
\begin{tabular}{lrrrrr}
\toprule
LangFlow arm & accept\,\% & \tailacc{} & $\Delta$\tailacc{} & $[2,3)$ & $z{<}2$ \\
\midrule
Human & 90.5 & 11.17 & --- & 2.41 & 0.74 \\
Stock & 60.5 & 8.28 & --- & 1.11 & 0.35 \\
Quota $(1,1,2)$ & 35.7 & 13.24 & $+4.96$ & --- & --- \\
Quota $(1,2,3)$ & 37.0 & 16.69 & $+8.41$ & 3.32 & 2.25 \\
\quad $+$ attestation gate & 38.0 & 15.05 & $+6.77$ & 4.10 & \textbf{0.87} \\
Quota $(2,3,4)$ & 39.0 & 24.16 & $+15.88$ & --- & --- \\
\bottomrule
\end{tabular}
\end{table}

\textbf{Reading the attestation audit.} The cost of the gate in Table~\ref{tab:appattest} depends on whether the model has spare qualifying positions to refill. On ELF-B, only the champion$+$co-pinning pair isolates the gate. Its gain moves from $+5.34$ to $+5.36$ raw, and from $+4.07$ to $+4.13$ after the foreign discount of Appendix~\ref{app:foreign}, both against the batch-matched stock of Table~\ref{tab:appdose} ($4.31$). Table~\ref{tab:appforeign} reads the same attestation-gated run, as its ELF-B \emph{fixed} row, against the batch-16 stock ($4.30$) and gives $+4.10$. The band-quota and dose-16 arms without the gate ran with an earlier German-only language gate that their gated counterparts drop, so the last column also removes that filter. After the foreign discount, their gains therefore fall ($+6.10\to+5.26$ and $+8.62\to+8.06$) as foreign-dominant words return, and the foreign share of the accepted-set tail rises from $14.5$ to $24.5\%$ and from $9.9$ to $14.8\%$. The LangFlow arms carry no language gate. As for provenance, the ELF-B champion$+$co-pinning commitment count and unattested share come from its batch-16 run ($6{,}220$ commitments) and the $\Delta$ columns from its batch-8 rerun ($6{,}219$), and the LangFlow rows are single runs on seed $42$. Over three seeds, the gated LangFlow fixed dose gains $+3.55$ (Table~\ref{tab:appforeign}).

\textbf{Reading the LangFlow quotas.} The LangFlow quota arms of Table~\ref{tab:applfquota} support two further readings. First, each commitment yields $2.5$--$6.8$ words in the targeted bands rather than one, so the overshoot is an amplification effect, not a matter of large quota numbers. Second, the three bands require very different doses to land on human: $0.39$, $1.09$, and $0.36$ of the commitments of the lightest quota, $(1,1,2)$, for $[3,4)$, $[2,3)$, and $z{<}2$, respectively. No uniform rescaling of a quota can therefore match the human profile on LangFlow. A per-band solve is required, and the linear fit that would give it is ill-conditioned at the three doses we ran.

\subsection{Reproduction reference}
\label{app:repro}

This subsection collects what is needed to rerun and re-score Table~\ref{tab:main}: the human reference values used across the paper, the per-seed values of Table~\ref{tab:main}, the sampler settings behind it, and the compute.

\textbf{Human reference values.} One human sample underlies every comparison in this paper, but it is read under several rules, so the same quantity takes different values in different tables. Table~\ref{tab:apphuman} lists these values, and a model reading is compared only with the human value under the same rule. Two probe-internal tails fit none of these rules, because they are read on the probes' own inputs: $7.454\%$ in the clean-embedding test (T5 tokens of $128$ documents) and $10.67$ at the start of the compounding probe (Appendix~\ref{app:exclusions}).

\begin{table}[ht]
\centering
\small
\renewcommand{\arraystretch}{0.92}
\caption{Human reference values and the rule behind each; the rules are defined in Appendix~\ref{app:metrics}.}
\label{tab:apphuman}
\begin{tabular}{>{\raggedright\arraybackslash}p{0.22\linewidth}>{\raggedright\arraybackslash}p{0.38\linewidth}>{\raggedright\arraybackslash}p{0.30\linewidth}}
\toprule
Value & Rule & Used in \\
\midrule
Tail $11.88\%$ & all-token, unfiltered, foreign discount on the numerator & Table~\ref{tab:main} \\
Tail $11.49\%$ & alphabetic, unfiltered, no discount & Table~\ref{tab:buckets} (three rare bands), Figure~\ref{fig:fig1}b, Tables~\ref{tab:apparms}, \ref{tab:appunfiltered}, \ref{tab:appllada}, \ref{tab:appsweep} \\
Tail $11.17\%$ ($11.2\%$) & alphabetic, accepted set under the dual bar (accept $90.5\%$) & every \tailacc{} reading, e.g.\ Tables~\ref{tab:appscale}, \ref{tab:appforeign}, \ref{tab:appaccept}, \ref{tab:appdose}, \ref{tab:applfquota} \\
Tail $12.35\%$ & top-5k tail, accepted set & Table~\ref{tab:appunfiltered} (right) \\
Non-words $1.14\%$ / $0.66\%$ & all-token / alphabetic, unfiltered & Table~\ref{tab:main} / Tables~\ref{tab:buckets}, \ref{tab:apparms}, \ref{tab:appbuckets} \\
$\zeta{=}0.971$; slope $-0.986$ & all-token; regular-expression tokens & Table~\ref{tab:main}; \S\ref{sec:collapse} \\
Types $17{,}726$ & all-token, mean of $20$ draws of $200$ documents & Table~\ref{tab:main} \\
Bars $2.737\%$, $10.12\%$ & per-sample 95th percentiles (non-words, 4-gram repetition) & every accept-then-measure reading \\
Clean-PPL $16.66$ & GPT-2-large, accepted set & Appendices~\ref{app:protocol}, \ref{app:accept} \\
\bottomrule
\end{tabular}
\end{table}

\textbf{Per-seed values.} Table~\ref{tab:appperseed} lists every seed of the multi-seed rows of Table~\ref{tab:main}, and Table~\ref{tab:apphyper} gives the settings of all its runs. The other rows of Table~\ref{tab:main} are single runs, except the ELF-B best-of-4 row, which selects among four stock runs. ELF-L is the one ELF model whose stock row also has three seeds. There, the tail and $|\Delta\zeta|$ keep their ordering on every seed, but JSD does not: the stock sampler reads $.0044$--$.0064$ and \graft{} $.0041$--$.0049$, so on seed $55555$ the stock reading is the lower one, and the JSD improvement of Table~\ref{tab:main} on ELF-L is within seed-to-seed variation. On LangFlow, the fixed-dose tail spans $1.18$--$1.33\times$ human across seeds, whereas the tuned dose reads $1.03\times$ on all three.

\begin{table}[ht]
\centering
\small
\renewcommand{\arraystretch}{0.92}
\caption{Per-seed values behind the three-seed rows of Table~\ref{tab:main}, scored as in that table ($n{=}200$ per seed). Their means agree with Table~\ref{tab:main} to within one unit in the last digit shown.}
\label{tab:appperseed}
\begin{tabular}{llrrrrrrr}
\toprule
Model & Row & seed & tail\,\% & $\times$human & non-word\,\% & JSD & $|\Delta\zeta|$ & types \\
\midrule
ELF-B & \graft{} & 31337 & 9.78 & 0.82 & 2.61 & .0049 & .138 & 13{,}757 \\
 &  & 55555 & 9.91 & 0.83 & 2.55 & .0047 & .111 & 14{,}037 \\
 &  & 77777 & 9.58 & 0.81 & 2.54 & .0052 & .141 & 13{,}693 \\
\addlinespace[2pt]
ELF-M & \graft{} & 31337 & 12.18 & 1.03 & 3.13 & .0082 & .061 & 15{,}832 \\
 &  & 55555 & 12.14 & 1.02 & 2.96 & .0083 & .063 & 15{,}853 \\
 &  & 77777 & 11.75 & 0.99 & 3.22 & .0072 & .075 & 15{,}400 \\
\addlinespace[2pt]
ELF-L & Stock & 31337 & 7.67 & 0.65 & 2.65 & .0049 & .068 & 15{,}419 \\
 &  & 55555 & 8.16 & 0.69 & 2.96 & .0044 & .040 & 15{,}909 \\
 &  & 77777 & 7.13 & 0.60 & 2.40 & .0064 & .067 & 15{,}107 \\
 & \graft{} & 31337 & 13.32 & 1.12 & 2.83 & .0049 & .011 & 17{,}001 \\
 &  & 55555 & 13.04 & 1.10 & 3.03 & .0048 & .005 & 17{,}304 \\
 &  & 77777 & 12.77 & 1.07 & 2.77 & .0041 & .013 & 17{,}014 \\
\addlinespace[2pt]
LangFlow & Stock & 42 & 9.03 & 0.76 & 3.23 & .0025 & .023 & 18{,}685 \\
 &  & 1042 & 9.25 & 0.78 & 3.24 & .0024 & .008 & 18{,}960 \\
 &  & 2042 & 9.19 & 0.77 & 3.35 & .0026 & .004 & 19{,}044 \\
 & \graft{} & 42 & 15.78 & 1.33 & 3.55 & .0075 & .058 & 19{,}103 \\
 &  & 1042 & 14.05 & 1.18 & 3.32 & .0044 & .064 & 19{,}046 \\
 &  & 2042 & 14.61 & 1.23 & 3.87 & .0055 & .068 & 18{,}794 \\
 & \graft{} (tuned) & 42 & 12.29 & 1.03 & 4.10 & .0001 & .040 & 18{,}754 \\
 &  & 1042 & 12.20 & 1.03 & 3.72 & .0003 & .034 & 19{,}139 \\
 &  & 2042 & 12.20 & 1.03 & 3.57 & .0002 & .060 & 18{,}928 \\
\bottomrule
\end{tabular}
\end{table}

\begin{table}[!b]
\centering
\small
\renewcommand{\arraystretch}{0.92}
\caption{Settings of the runs in Table~\ref{tab:main}, as read from the arguments stored with each run, for the procedure of Algorithm~\ref{alg:graft}. A cell that spans both columns applies to both model families. The ELF-B tuned entries describe the arm of Appendix~\ref{app:heads}, and the LangFlow best-of-4 seed an arm that \S\ref{sec:controls} quotes; neither is a row of Table~\ref{tab:main}.}
\label{tab:apphyper}
\begin{tabular}{>{\raggedright\arraybackslash}p{0.23\linewidth}>{\raggedright\arraybackslash}p{0.38\linewidth}>{\raggedright\arraybackslash}p{0.27\linewidth}}
\toprule
Setting & ELF-B / M / L & LangFlow \\
\midrule
Checkpoint, vocabulary & official OpenWebText (105M / 342M / 652M), T5 ($32$k) & official OpenWebText (652M), GPT-2 BPE \\
Sampler & SDE, $32$ steps, churn $\gamma{=}1.5$, self-conditioning guidance $3$, logit-normal schedule$^{a}$ & deterministic Euler, $256$ steps, linear time grid \\
Length; samples, batch & $1{,}024$; $200$, batch $16$ (ELF-B tuned: $8$) & $1{,}024$; $200$, batch $8$ (stock, seed $42$: $16$) \\
Seeds; read time & $31337$, $55555$, $77777$ (best-of-4 adds $99999$); $t_{\mathrm{read}}{=}0.20$ & $42$, $1042$, $2042$ (best-of-4 adds $3042$); $0.72$ \\
Selection & \multicolumn{2}{p{0.65\linewidth}}{$R{=}4$, $m{=}8$; candidates among the $512$ most likely words, rare words $0<z<4$; $q_{\min}{=}0.08$; spacing $s{=}4$; atomic commitment, splicing and co-pinning on} \\
Attestation gate & \multicolumn{2}{p{0.65\linewidth}}{on; attestation lexicon of documents $1{,}000$--$30{,}999$ of the reference shard, minimum count $1$} \\
Language gate & on ($13$ languages, margin $2.5$) & off \\
Anti-repetition correction & $\lambda{=}0.5$; direction from ELF-B, unit norm & $\lambda{=}0.5$; own, unnormalized \\
Tuned dose & quota $(8,5,3)$, $q_{\min}{=}0.02$ (ELF-B, Appendix~\ref{app:heads}) & quota $(0,1,1)$, $q_{\min}{=}0.02$ \\
\bottomrule
\multicolumn{3}{p{0.90\linewidth}}{\footnotesize $^{a}$Center $-1.5$, or $-1.0$ (the step-schedule shift) for the ELF-B \graft{} and argmax-control rows.}
\end{tabular}
\end{table}

\textbf{Compute.} Every run fits on one 24\,GB NVIDIA RTX 3090. The experiments ran on two machines with three such GPUs each, one arm per GPU. With model loading included, an arm of $n{=}200$ samples takes about $1$--$3$ minutes (stock) and $2$--$3$ minutes (\graft{}) on ELF-B, about $10$ minutes for \graft{} on ELF-M and $15$ on ELF-L, and about $20$ minutes (stock) and $40$ minutes (\graft{}) on LangFlow. Each ELF-S arm of Appendix~\ref{app:training} trained for about $8$ hours.

\clearpage
\section{Extended related work}
\label{app:related}

This appendix extends \S\ref{sec:related} to the rest of both lines and to nearby work on where a model draws a discrete word.

\textbf{The continuous line.} Diffusion-LM put Gaussian diffusion on learned word embeddings and had to round them back to the vocabulary at every step \citep{li2022diffusionlm}. Analog Bits added self-conditioning \citep{chen2023analog}, CDCD predicted the clean embedding at every step as the average of token embeddings weighted by a predicted distribution over the vocabulary \citep{dieleman2022cdcd}, and Difformer addressed a collapsing embedding space and insufficient noise \citep{gao2024difformer}. \citet{shabalin2026whyfail} traced the failures of Gaussian diffusion on discrete data to a sampling interval in which the noised density is multimodal. Latent-space variants diffused in the latent space of a separately trained autoencoder and left the return to tokens to its decoder \citep{lovelace2023latent}, and \citet{gulrajani2023plaid} found that their likelihood-trained Plaid models need about $64\times$ the compute of autoregressive models to reach the same likelihood. The recent revival, comprising ELF \citep{hu2026elf}, LangFlow \citep{chen2026langflow}, flow map language models \citep{fmlm2026}, and CoDAR \citep{codar2026}, rounds to the vocabulary at no intermediate step, and ELF goes furthest by supervising tokens only at the final step. Yet the per-step anchors that this line removed, rounding among them, were also its only place to draw a word (\S\ref{sec:related}). ConvergeFlow \citep{convergeflow2026} goes further, dropping the decoding head trained by cross-entropy and confining an MSE-trained prediction to the convex hull of the token embeddings. Because that prediction is still a conditional mean, our mechanism predicts that ConvergeFlow loses the tail as well, a test we have not run.

\textbf{The discrete line.} Discrete models have the sampling interface natively: from D3PM's absorbing-state corruption \citep{austin2021d3pm} through SEDD \citep{lou2024sedd} to the simplified masked models \citep{sahoo2024mdlm,shi2024md4}, every unmasking step draws tokens from a categorical distribution. \citet{ou2024radd} proved masked-diffusion training equivalent to any-order autoregression, LLaDA scaled the recipe to 8B parameters \citep{nie2025llada} and beyond \citep{llada2}, and block diffusion runs diffusion within blocks that are generated in order \citep{arriola2025bd3}. Because these models sample, a fault can also sit in the sampler itself: \citet{zheng2025masked} showed that float32 truncation of Gumbel noise silently lowers the sampling temperature, and with it the diversity of the text. That fault shares the symptom of our defect but has a one-line numerical cause, whereas ours is structural; Appendix~\ref{app:checks} checks the \graft{} sampler for it.

\textbf{When positions are decided.} For discrete parallel decoding, \citet{feng2025theory} separate a perplexity regime, where a constant number of steps suffices, from a sequence-correctness regime, where steps must scale with length; \citet{kim2025train} show that an adaptive unmasking order sidesteps a masked model's hardest subproblems; and \citet{ctc2026} prove that conditional total correlation is the exact cost of revealing positions in parallel. Work on when a diffusion trajectory settles a feature, including critical windows \citep{li2024critical} and the entropic signature of speciation \citep{jiang2026speciation}, supplies the tools behind our read-window measurements (Appendix~\ref{app:commitlog}). Our frequency-resolved finding, that rare words are decided later and over a wider span than common ones while ELF's step schedule is centered earlier, appears to be new for continuous dLLMs.

\textbf{Why the denoising head averages.} ELF shares three choices with recent image generation: continuous rather than vector-quantized tokens \citep{li2024mar}, x-prediction for stability in high dimensions \citep{li2025jit}, and MeanFlow-style training-time classifier-free guidance \citep{geng2025meanflow}. The second choice underlies the split that our diagnosis exploits, between a denoising head trained by MSE, which regresses to the conditional mean, and a decoding head trained by CE, which spreads probability over every plausible word. Under MSE, x-prediction \emph{is} a conditional-mean regression, which \citet{blau2018perception} warn cannot also preserve the data distribution.

\textbf{The nearest samplers.} MCB \citep{mcb2026} and INCO \citep{inco2026} are the two samplers closest to \graft{} among those proposed for continuous-space language models. MCB makes a closely related argument for flow language models over one-hot tokens---that a standard sampler collapses each position's distribution over clean tokens to its conditional mean---and therefore samples a clean endpoint at every step and bridges toward it. INCO instead redistributes denoising progress by inverse confidence and never draws a token. Sticky jump diffusions \citep{sticky2026} derive when a continuous state should jump onto a token embedding. \graft{} shares the premise of MCB but neither samples nor selects at every step: it selects once per round, at the read time, commits the strongest losing rare candidate, and holds that commitment from the start of the next trajectory. At an equal number of function evaluations on LangFlow, neither MCB nor INCO raises the tail (Appendix~\ref{app:heads}).

\textbf{The repetition defect.} \citet{lpir2026} diagnosed ELF's \emph{repetition} defect as a self-conditioning feedback loop, located through a single repetition direction and corrected by a small subtraction at every step, whose usable strength ends at an off-manifold breakdown. Our defect lies on a different axis: turning that feedback from off to full leaves the tail flat (Table~\ref{tab:appexcl}). Their own appendix likewise treats non-words as a second defect, independent of repetition. From them we take the method, namely acceptance bars calibrated on human text, direction estimation with placebo controls, keeping the correction strength within its usable range, and transfer to an independent implementation. We add a mechanism, the conditional-mean readout with a rule that predicts which rare words it loses; the finding that, in our experiments, only words given before the first step persist reliably; and a sampler built from both. \graft{} includes their anti-repetition correction because the repetition that commitments add matches that feedback (\S\ref{sec:method}); the two defects thus meet in the method, not in the mechanism.

\textbf{Adjacent work.} DSL-LLaDA \citep{dsl2026} moves in the direction opposite to ours, adapting the discrete LLaDA to continuous denoising that defers the choice of a token to the final step. Together, the two directions are complementary evidence that each line lacks what the other has. Untied self-conditioning \citep{untied2026} documents a train--inference mismatch in self-conditioning, which motivated the variant of Appendix~\ref{app:negative} that leaves the self-conditioning input unfixed at committed positions. \citet{lost2026} find embedding norms flat across frequency rank in a masked diffusion model, in line with our exclusion of a geometric amplifier in T5-small's static embedding table (Appendix~\ref{app:exclusions}), although model and statistic differ. On evaluation, LongTail-Swap \citep{longtailswap2025} benchmarks rare-word learning in models trained on small corpora, and \citet{ardiversity2025} find that chat-tuned autoregressive models differ from human writers in lexical diversity. On unfiltered output, a decoding rule alone moves GPT-2 from $0.00$ to $1.27\times$ the human tail (Table~\ref{tab:apparms}).

\section{Qualitative samples}
\label{app:qualitative}

This appendix shows excerpts of unconditional output so that the committed words can be judged in context, and it closes with examples of fabricated names. On ELF-B and LangFlow, the champion$+$co-pinning arm, an earlier configuration without gates (Appendix~\ref{app:configs}), appears beside the stock sampler. A fixed-seed random number generator (seed~0) drew the samples from the \emph{accepted} set of each arm; in the champion$+$co-pinning arms, we also required at least four kept commitments so that each excerpt can display some. A champion$+$co-pinning excerpt is cut from the densest 55-word span around commitments, located by script rather than by hand, and shortened with ellipses for display, with the committed words in \textbf{bold}; a stock excerpt is the opening of its sample. One drawn ELF sample (index~10) describes graphic violence, so we exclude it from display and disclose the exclusion rather than draw a replacement. Full outputs for every arm will be released with the code.

\textbf{ELF-B, champion$+$co-pinning} (sample 139): \emph{``short passes, and create space between areas where a goal should usually \textbf{concede}. During that time, he provided a \textbf{viewer} of moments that would have come to be impossible for Chelsea's young \textbf{talents}, including Jos\'e Benis of Real-Madrid, David Silva of Manchester United, and now Alexis Edgelen of the \textbf{Croatia} Firm. Peterin's performance \textbf{devastated}\,\ldots''}
(sample 155): \emph{``are having their first \textbf{triumph} of the season since a 1-1 \textbf{clash} at the Germanpark Stadium\,\ldots\ \textbf{Kiev} have also lost their first qualifying match since a 2-1 defeat \textbf{Tottenham} Tra refford on Dec.\ 20. Top position competitors Manchester United, \textbf{sus} are currently No.\ 10 in the \textbf{Bundesliga} League\,\ldots''}
Most commitments are real rare words placed plausibly, such as \emph{concede}, \emph{triumph}, \emph{Bundesliga}, and \emph{Kiev}. Fabricated names persist around them (\emph{Edgelen}, \emph{Peterin}), but these are the stock sampler's pathology (\S\ref{sec:collapse}), not something that committing introduced. A few commitments are fragments that have a lexicon entry, or foreign-dominant words (\emph{sus}, German function words). The fragments are the sub-word cost of committing measured in Appendix~\ref{app:commitlog}; the foreign-dominant words appear because the arm is ungated, and the language gate removes them (Appendix~\ref{app:foreign}).

\textbf{ELF-B, stock} (sample 93): \emph{``In response to a series of violent car accidents in several neighborhoods, the United States.\ government has launched a new form of technology to try to improve the names of violent individuals\,\ldots''} (sample 196): \emph{``Using a standard pattern of V2 colors results in a range of V2 colors increasing. As the V2 standard increases, the range of V2 colors from V2 to 5, can be completely reverseed\,\ldots''} Both excerpts read fluently, but they use generic vocabulary and loop in a way that borders on repetition.

\textbf{LangFlow, champion$+$co-pinning} (sample 98): \emph{``\textbf{WikiLeaks} would be prepared to accompany the May 4 Olympic Ceremony, under which will be met at the Russian capital, to Sochi. `I made the statement that reps from the Russian government are aware that I will not play for Sochi by May of this year,' \textbf{Podesta} said in a statement issued to \textbf{Politico}\,\ldots''} (sample 76): \emph{``\textbf{Romney} presidential bus 7 Arab \textbf{sidewalk} ahead as of Riyadh 7 Washington Co 7 Washington Monument looking near Riyadh\,\ldots''} The second excerpt reads like a run of photo captions and headlines. It passes the acceptance bars because those bars are calibrated on human OpenWebText, which contains this register as well.

\textbf{LangFlow, stock} (sample 172): \emph{``concept.'\ A Xbox 360 exclusive announced last November and that the Xbox 360 listing could link up with Xbox One Now\,\ldots''} (sample 63): \emph{``Leaks has been releasing leaks that rely on a combination of conspiracy theories. Nothing Assange does not want to provoke the leak is that Julian Assange released a file of WikiLeaks documents\,\ldots''}

\textbf{Fabricated-name exhibits} (the diagnosis of \S\ref{sec:collapse}). In these exhibits, bold marks fabricated names rather than commitments. The XSum-conditioned ELF-B checkpoint writes, verbatim: \emph{``Two men and a boy have been charged over a aggravated vehicle taking at \textbf{Portakta}, near \textbf{Goldfed}, in west Belfast''}; \emph{``New Barcelona president \textbf{Neyll} has resigned from the Spanish club''}. Unconditional LangFlow output contains non-words such as \emph{Morians}, \emph{CARDERON}, \emph{Wexman}, \emph{Villarue}, and \emph{kudawa}. Across the step counts of Appendix~\ref{app:sweep}, $60$--$69\%$ of its non-words are capitalized, the signature of fabricated names in proper-noun positions that both models share.

\end{document}